\documentclass[letterpaper,twocolumn,10pt]{article}
\usepackage{usenix,endnotes}

\usepackage{xspace}
\usepackage{enumitem}
\usepackage{subcaption}
\usepackage{booktabs}
\usepackage{multirow}
\usepackage{xcolor}
\usepackage{graphicx}
\usepackage{amsmath}
\usepackage{amssymb}
\usepackage{bm}
\usepackage[ruled,linesnumbered]{algorithm2e}

\newcommand{\SysName}{\textsf{ZeRO-Prefill}\xspace}
\newcommand{\SysBackend}{AsyncEP\xspace}
\newcommand{\SysBackendName}{Asynchronous Expert Parallelism}

\title{{\SysName}: Zero Redundancy Overheads in MoE Prefill Serving} 
\author{
{\rm Zhaoyuan Su$^{1,2}$\quad Olatunji Ruwase$^{2}$\quad Karthik Ganesan$^{2}$\quad Aurick Qiao$^{2}$}\\
{\rm Samyam Rajbhandari$^{2}$\quad Juncheng Yang$^{3}$\quad Yue Cheng$^{1}$\quad Yuxiong He$^{2}$}\\[2pt]
$^{1}$University of Virginia\qquad $^{2}$Snowflake AI Research\qquad $^{3}$Harvard University
}
\date{}

\begin{document}

\maketitle
\begin{abstract}

Production LLM workloads increasingly serve discriminative tasks, such as classification, recommendation, and verification, whose answers are read from the logits of a single prefill pass with no autoregressive decoding.
Serving these \emph{prefill-only workloads} on mixture-of-experts (MoE) models is bottlenecked not by compute but by the distributed execution required to fit the model: existing parallel strategies (tensor, expert, and pipeline parallelism) trade memory pressure for redundant computation, communication, and synchronization, severely degrading MoE prefill serving efficiency.

We observe that these overheads stem from coupling expert placement with synchronous activation routing---a design inherited from the decoding era.
The long, compute-bound forward passes of large-batch prefill open a per-layer window wide enough to stream expert weights in the background, replacing per-layer activation \texttt{AllToAll} with asynchronous weight \texttt{AllGather} fully overlapped with computation.
We propose {\SysName}, a prefill-only serving system built on this idea. Its backend, {\SysBackend} (\textit{\SysBackendName}), gathers experts \emph{by weight} rather than \emph{by activation}; its co-designed frontend enforces a physically derived saturation threshold through prefix-aware routing and true-FLOPs load tracking.
On Qwen3-235B-A22B across four hardware/precision configurations, \SysName\ delivers {1.34--1.37$\times$} throughput over the strongest distributed baseline on real-world workloads and up to {1.59$\times$} on long-context synthetic workloads, sustaining {29.8--36.2\%} per-GPU model FLOPs utilization.

\end{abstract}
\section{Introduction}
\label{sec:intro}

\begin{figure}[t]
    \centering
    \includegraphics[width=\linewidth]
    {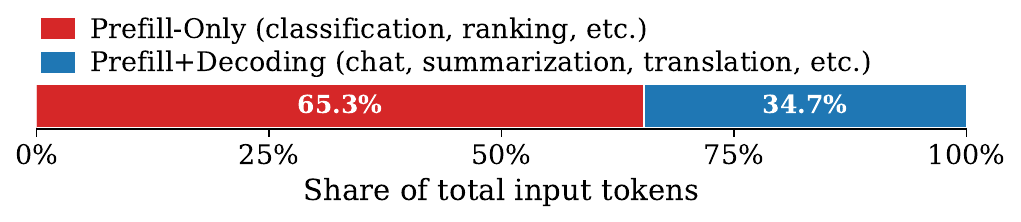}
    \caption{Prefill-only workloads dominate LLM serving input-token traffic in a production system.}
    \label{fig:motivation_prefill_share}
    \vspace{-15pt}
\end{figure}

Large language models (LLMs) are increasingly used beyond text generation for a class of discriminative tasks---classification~\cite{sun2023text, qin2023chatgpt}, moderation~\cite{inan2023llama}, recommendation~\cite{zhang2026recommendation}, factual verification~\cite{min2023factscore}---where the answer is one choice from a small, predefined candidate set and is fully determined by the logits of a single forward pass, the pattern behind emerging decision models such as Jev~\cite{jev} (Appendix~\ref{app:jev}). Measurements from a production cluster (Fig.~\ref{fig:motivation_prefill_share}) show that such \emph{prefill-only workloads} already account for {65.3\%} of all input tokens served. Their system profile differs from generative serving along three axes: they are throughput-oriented rather than latency-sensitive, run at extremely large batch sizes that yield long compute-bound forward passes, and exhibit abundant prefix sharing (shared system prompts and user profiles) both within and across batches---an optimization regime that existing decoding-oriented serving systems cannot exploit. PrefillOnly~\cite{prefillonly} optimizes KV-cache lifetime and job-level scheduling for models that fit on a single GPU, whereas \SysName\ targets large MoE models requiring distributed serving; we use PrefillOnly as a baseline (\S\ref{sec:evaluation}).

Meanwhile, state-of-the-art open-weight LLMs have shifted to the Mixture-of-Experts (MoE) architecture~\cite{cai2025survey, shazeer2017outrageously}, whose total parameter counts grow far faster than per-GPU HBM (\S\ref{subsection:mem_pressure}) and force distributed execution purely to hold the weights. Distributed parallelism relieves memory pressure only by introducing \emph{three redundancies} (\S\ref{sec:analysis_system}) on the MoE prefill critical path: \emph{redundant computation} from sub-saturated per-device GEMMs and routing-imbalance stragglers, \emph{redundant memory} from full-resident expert weights, $k$-way activation expansion, and duplicated prefix KV caches, and \emph{redundant communication} from two synchronous \texttt{AllToAll}s per MoE layer under expert parallelism. 

These redundancies are not inherent to MoE serving---they are inherited from a decoding-era design choice that \emph{couples expert placement with synchronous activation routing}. The long, compute-bound forward passes of large-batch prefill open a wide per-layer compute window; if expert weights are streamed into the window in the background, per-layer \texttt{AllToAll} can be removed from the critical path entirely. Concretely, experts are gathered \emph{by weight} rather than routed \emph{by activation}, so every GPU holds the complete expert set for the current layer, and dispatch becomes a local operation.

We propose {\SysName}, a prefill-only serving system that eliminates all three redundancies on the MoE prefill critical path. Its core execution paradigm {\SysBackend} (\SysBackendName) replaces per-layer activation \texttt{AllToAll} with background expert-weight \texttt{AllGather}, overlapped with computation and optionally extended to CPU-DRAM offloading. The overlap is governed by a \emph{saturation threshold} $T$ (\S\ref{subsec:async_ep_d2d}) derived from hardware, workload, and model configuration. The \SysName\ frontend enables $T$ as a per-GPU batching rule and, on top of it, performs prefix-aware routing and true-FLOPs compute tracking, closing the frontend--backend co-design loop.

We implement \SysName\ on top of vLLM and evaluate it across four hardware/precision combinations (8$\times$A100 BF16, 8$\times$H100 BF16/FP8, 8$\times$H200 FP8) against the five distributed MoE strategies in vLLM and four PrefillOnly configurations. On an aggregated real-world prefill-only workload, \SysName\ achieves {1.34--1.37$\times$} end-to-end throughput over the strongest baseline in every hardware/precision/parallel-degree cell, and up to {1.59$\times$} on long-context synthetic workloads. Its deployment envelope widens from ``$\geq$4 GPUs required to hold Qwen3-235B-A22B'' to ``1--8 GPUs'' (a $4\times$ broader hardware range), while sustaining {29.8--36.2\%} per-GPU model FLOPs utilization (MFU). 

This paper makes four contributions:
\begin{itemize}[nosep,leftmargin=*]
  \item A workload characterization formalizing the atomic \emph{prefill-as-a-service} operation and consolidating three structural opportunities of prefill-only serving (\S\ref{sec:analysis_workload}).
  \item A quantitative analysis of MoE prefill bottlenecks unified in a per-device per-layer communication table over eight parallel-strategy combinations (\S\ref{sec:analysis_system}).
  \item \SysBackend, an MoE execution paradigm that gathers experts \emph{by weight} rather than \emph{by activation}, removing per-layer \texttt{AllToAll} and routing-imbalance stragglers from the critical path (\S\ref{sec:backend}).
  \item \SysName, a co-designed serving system bound by a physically derived saturation threshold, delivering $1.34$--$1.37\times$ throughput and a $4\times$ broader deployable envelope on large MoE models (\S\ref{sec:frontend}--\S\ref{sec:evaluation}).
\end{itemize}

\section{Background}
\label{Sec:background}

\subsection{LLM Inference: Prefill and Decoding}
\label{subsec:bg_inference}
LLM serving executes in two phases: \emph{prefill} processes the entire input context of length $S$ in a single forward pass, while \emph{decoding} autoregressively emits output tokens one at a time. To amortize attention across decoding steps, engines such as vLLM~\cite{vllm} and SGLang~\cite{sglang} cache per-layer key/value projections, whose footprint scales linearly with batch size and sequence length (full memory accounting in \S\ref{subsection:mem_pressure}).

\subsection{Prefill-Only LLM Workloads}
\label{subsec:bg_prefill_only}
A growing class of production tasks---classification, moderation, recommendation, verification---can be answered from the logits of a single prefill pass, with no autoregressive decoding. \SysName\ targets this setting exclusively, with one output token per request. We characterize these \emph{prefill-only workloads} in \S\ref{sec:analysis_workload}.

\subsection{Mixture-of-Experts (MoE) Models}
\label{subsec:bg_moe}
Most recent open-weight LLMs---Qwen3~\cite{qwen3}, DeepSeek-V3~\cite{deepseekv3}, GLM-4.5~\cite{glm45}, Mixtral~\cite{mixtral}, gpt-oss~\cite{gptoss}---adopt the {Mixture-of-Experts (MoE)} architecture to scale capacity without proportional growth in per-token FLOPs. In an MoE Transformer, the dense FFN of (some or all) decoder blocks is replaced by $E$ parallel experts, each a two-layer MLP; a \emph{router} dispatches each token to its top-$k$ experts (typically $k{=}2$--$8$) and sums their outputs. Sparse activation packs hundreds of billions of total parameters while activating only a small subset per token, but the cost is paid on the \emph{system} side: total parameters far exceed per-GPU HBM, and top-$k$ routing is typically imbalanced across experts and GPUs.

A structural feature of MoE is that {attention weights and expert weights are disjoint parameter groups}. The two stacks can therefore adopt \emph{independent} parallelization strategies; this decoupling makes combinations such as DP+EP or TP+EP meaningful and is exploited by our design in \S\ref{sec:backend}.

\subsection{Distributed Parallelism for MoE Serving}
\label{subsec:bg_parallelism}
Because large MoE models rarely fit on a single GPU, serving systems resort to distributed execution. Five strategies are in common use: DP~\cite{li2020pytorch_DP} (replicate weights, shard requests), TP~\cite{shoeybi2019megatron} (shard weight matrices), PP~\cite{huang2019gpipe, narayanan2019pipedream} (pipeline layers), SP~\cite{liu2023ring_SP, korthikanti2023reducing} (shard along sequence), and the MoE-specific EP~\cite{lepikhin2020gshard_EP} (shard experts), each with distinct on-path communication semantics. Because attention and expert stacks are disjoint (\S\ref{subsec:bg_moe}), MoE systems can compose different parallel strategies for the two stacks; \S\ref{sec:analysis_system} (Table~\ref{tab:moe_comm_compare}) gives a per-device per-layer comparison of the relevant combinations.

\section{Workload Analysis}
\label{sec:analysis_workload}

\subsection{Prefill-Only Workloads in Practice}
\label{subsection:prefill-only_in_practice}

\noindent\textbf{Definition and atomic operation.}
\emph{Prefill-only workloads} are inference tasks whose output is selected from a predefined candidate token set via the logits of a single prefill forward pass, without autoregressive decoding. We formalize the atomic \emph{prefill-as-a-service} operation: given an input context $c$ and a candidate token set $\mathcal{C} = \{t_1, \dots, t_{m}\}$, the system performs one prefill pass and returns $t^{*} = \arg\max_{t\in\mathcal{C}} \text{logit}(t \mid c)$. Unlike autoregressive serving, this operation has no iterative decoding and no per-token KV updates, making it inherently parallelizable and throughput-friendly.

\noindent\textbf{Representative categories.}
A broad range of industrial tasks reduces to this atomic operation, directly or via reformulation:
\begin{itemize}[noitemsep,leftmargin=*]
\item \emph{Single-Token Classification} (binary sentiment, toxicity, factual verification): the candidate set is a small set of single-token labels, served by one atomic call.
\item \emph{Single-Choice Selection} (multiple-choice QA, intent classification): assigning each candidate a symbolic identifier (A/B/C/D) in the prompt reduces the task to selecting one identifier token---again one atomic call.
\item \emph{Multi-Selection} (multi-label classification, recommendation, ranking): we \emph{decompose} one $N$-candidate request into $N$ independent binary sibling requests, served by $N$ parallel atomic operations. The $N$ siblings share the same prefix; prefix sharing (\S\ref{subsection:prefix_sharing}) absorbs the duplication, keeping total prefill compute comparable to a single undecomposed pass while eliminating decoding entirely.
\end{itemize}

\noindent\emph{\underline{Observation \#1.} As prior work also observes~\cite{prefillonly, zheng2024batchllm}, a wide class of production LLM workloads can be served through prefill-as-a-service, motivating serving systems specialized for prefill-only execution.}

\subsection{Workload Characteristics}

Unlike latency-sensitive interactive serving, prefill-only workloads are \emph{throughput-oriented}: the primary objective is to minimize total time over a large batch, not per-request latency. These workloads run in offline pipelines~\cite{zheng2024batchllm, vellaisamy2025characterizing} that process millions to billions of requests in bulk, with input lengths spanning short reviews to long documents of tens of thousands of tokens. Each prefill forward therefore processes an enormous token count and runs in a heavily compute-bound interval, fundamentally different from the memory-bandwidth-bound steps of autoregressive decoding. Such a long compute window can hide \emph{weight streaming}: loading upcoming layers' weights from CPU memory or peer GPUs while the current layer computes (\S\ref{sec:backend}).

\noindent\emph{\underline{Observation \#2.} The long compute-bound forward passes of large-batch prefill open an opportunity to overlap weight streaming with computation, motivating weight offloading~\cite{flexgen, moelightning} as a first-class design choice that unlocks MoE models larger than GPU memory.}

\subsection{Prefix Sharing Opportunities}
\label{subsection:prefix_sharing}

Prefix reuse appears at two scales. \emph{In-batch}, multi-selection decomposition (\S\ref{subsection:prefill-only_in_practice}) and similar reformulations produce batches in which many requests share a common prefix (e.g., $N$ candidate-scoring requests all prefixed by the same user profile); since attention is causal, the prefix attention output is identical across siblings, so computing it once eliminates nearly all prefix-stage compute and collapses $N$ redundant copies of the prefix KV into a single HBM copy. \emph{Cross-batch}, production pipelines process requests built from shared templates, stable user profiles, or document headers (e.g., the same user profile scored against successive candidate sets); caching the prefix KV across batches skips both recomputation and memory-bandwidth cost for subsequent same-prefix requests, and the cached prefix occupies HBM only once.

\noindent\emph{\underline{Observation \#3.} Prefill-only workloads exhibit abundant prefix sharing, both within and across batches, motivating prefix-aware scheduling and caching as a design principle.}

Observations~\#1--\#3 are opportunities that existing serving systems fail to capture on large MoE models, as \S\ref{sec:analysis_system} shows.
\section{System Analysis}
\label{sec:analysis_system}

Under prevailing designs, prefill-only MoE serving suffers a causal chain of three inefficiencies: \emph{memory pressure forces distributed parallel execution, which in turn degrades compute efficiency and introduces heavy on-path communication}. Throughout, we use the following notation: $B$~batch size, $S$~sequence length, $L$~layers, $H$~hidden size, $N_{kv}$~KV heads, $d_h$~head dim, $h$~MoE intermediate size, $k$~top-$k$ routing fan-out, $b$~bytes/element (BF16: $b{=}2$, FP8: $b{=}1$).

\subsection{Memory Pressure}
\label{subsection:mem_pressure}

\begin{figure}[t]
  \centering
  \includegraphics[width=\linewidth]{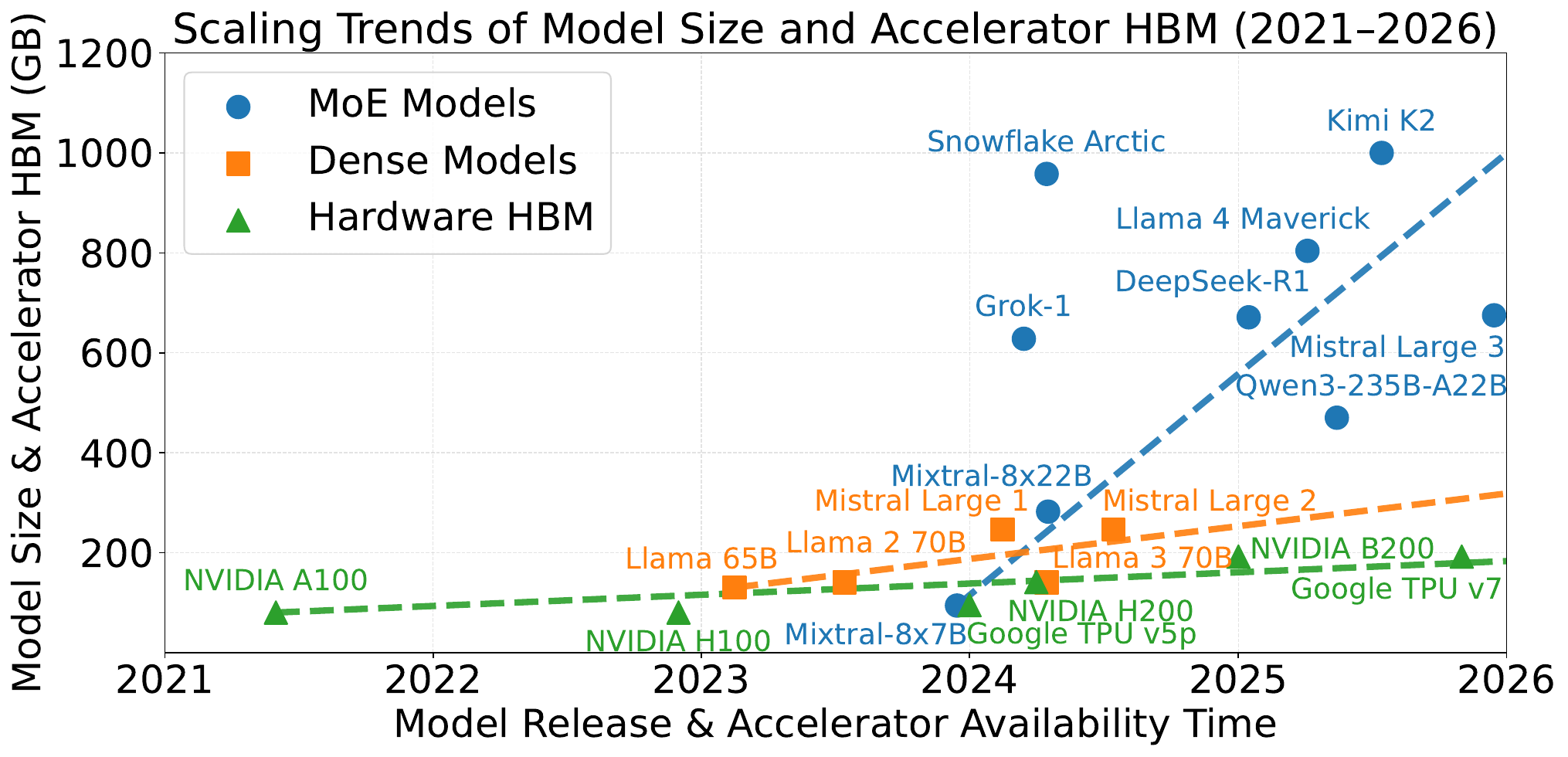}
  \caption{MoE model size vs.\ per-GPU HBM capacity across hardware generations. \textit{Total parameters grow far faster than HBM; the structural gap persists even under FP8.}}
  \label{fig:model_hardware_trend}
  \vspace{-15pt}
\end{figure}

Three components contend for GPU HBM during prefill-only MoE serving:

\noindent\textbf{Model weights.}
MoE scales by adding experts and widening expert MLPs, so total parameters grow far faster than per-GPU HBM capacity across hardware generations (Figure~\ref{fig:model_hardware_trend}). The gap persists even under FP8, placing moderately sized MoE deployments beyond the reach of a single GPU.

\noindent\textbf{KV caches.}
$V_{\text{KV}} = 2 L B S \, N_{kv} d_h \, b$, linear in both $B$ and $S$. Under the long-context, large-batch regime of \S\ref{sec:analysis_workload}, KV grows without bound with batch and sequence length, rapidly overtaking weights to become the dominant HBM consumer.

\noindent\textbf{Activations.}
Approximating peak activation by the two largest MLP intermediates, $V_{\text{act}} \approx 2\,B S\, k\, h\, b$---linear in $B{\cdot}S$ and amplified by a factor of $k$ from top-$k$ routing, unlike dense models where activations traverse a single MLP path. Under aggressive batching and wide MoE, activations themselves consume a non-trivial slice of HBM.

\noindent\emph{\underline{Insight \#1.} Large-MoE prefill serving faces severe HBM pressure from weights, KV caches, and activations simultaneously, forcing existing systems into distributed parallel execution.} 

\subsection{Computation Inefficiency}
\label{subsection:compute_inefficiency}

Once distributed execution is chosen, compute efficiency degrades at two levels: insufficient per-device token batches (model level) and MoE routing imbalance (layer level).

\noindent\textbf{Model level: MFU  requires large token batches.}
Model FLOPs Utilization~\cite{chowdhery2023palm_MFU} ($\text{Achieved FLOPs}/\text{Peak FLOPs}$) is primarily determined by the effective token batch $B{\times}S$ and the resulting GEMM geometry---larger batches improve arithmetic intensity and Tensor Core occupancy.

\begin{figure}[h]
  \centering
  \includegraphics[width=\linewidth]{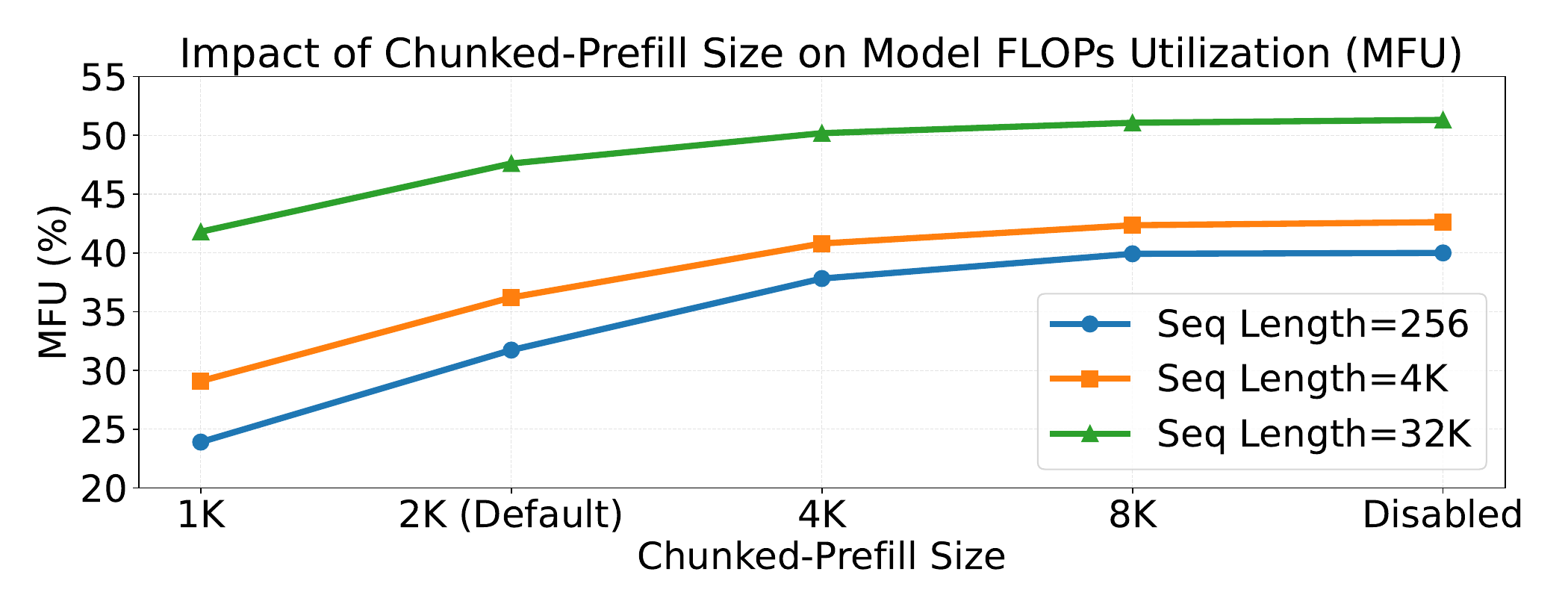}
  \caption{BF16 MFU of Qwen3-30B-A3B on A100 (80GB, vLLM) across chunked-prefill~\cite{sarathi} sizes and sequence lengths ($B{\times}S{=}32$K tokens, output 1, pure prefill).}
  \label{fig:mfu_chunked_prefilling}
  \vspace{-5pt}
\end{figure}

Figure~\ref{fig:mfu_chunked_prefilling} shows that small chunks (2K, by default on A100) constrain the per-kernel token batch and reduce GEMM dimensions; larger or disabled chunking recovers MFU, and longer contexts further improve it by enlarging per-request token volume. In short, \emph{MFU degrades whenever distributed execution or aggressive chunking drives per-device token batches below the GEMM-saturation threshold.}

\noindent\textbf{Layer level: expert-load imbalance fragments GEMMs.}
Even with a large global batch, MoE top-$k$ routing distributes tokens unevenly across experts~\cite{zhou2022mixture, stmoe}.

\begin{figure}[t]
  \centering
  \begin{subfigure}{\linewidth}
    \centering
    \includegraphics[width=\linewidth]{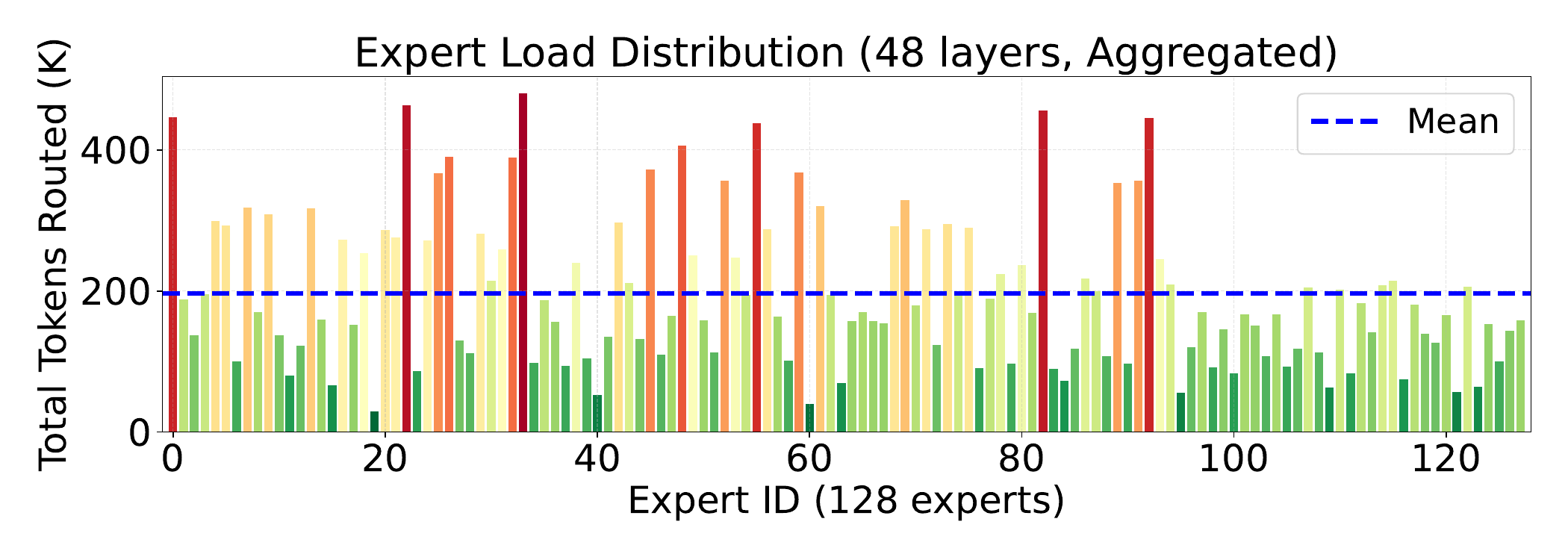}
    \caption{Aggregated expert load across all 48 layers}
    \label{fig:expert_imbalance_aggregated}
  \end{subfigure}
  \begin{subfigure}{\linewidth}
    \centering
    \includegraphics[width=\linewidth]{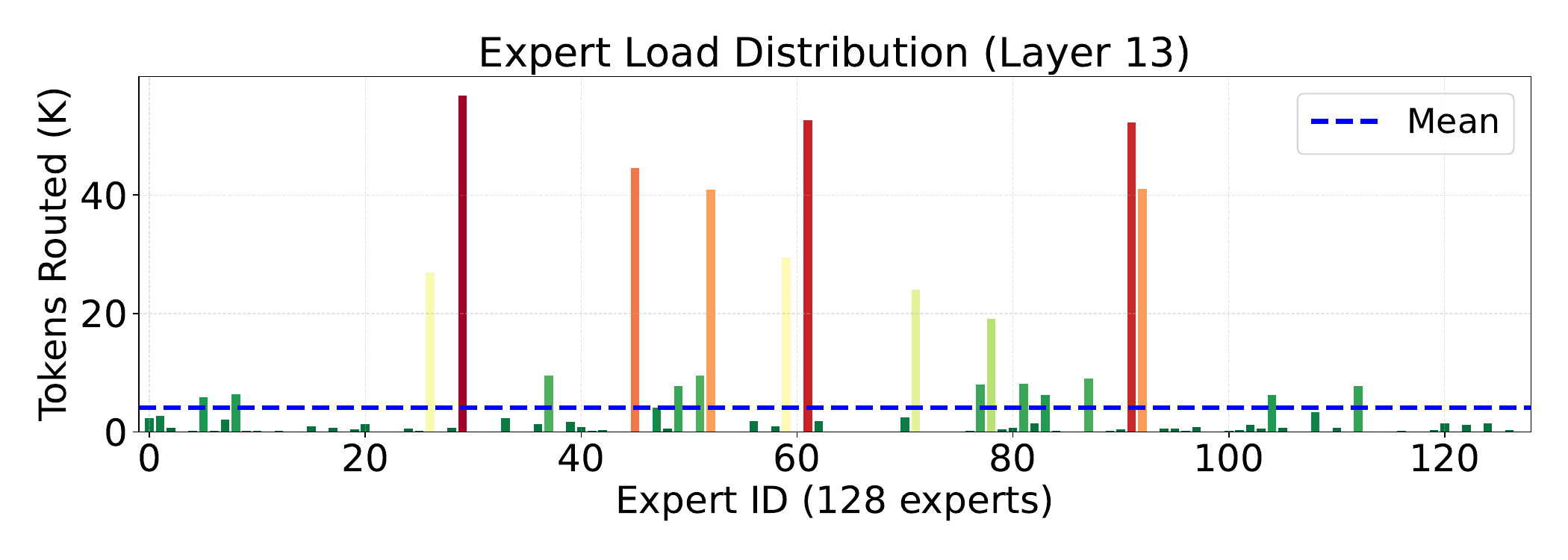}
    \caption{Expert load distribution in a single MoE layer (layer 13)}
    \label{fig:expert_imbalance_layer_13}
  \end{subfigure}
  \caption{Expert routing imbalance of Qwen3-30B-A3B on A100 (vLLM, BF16) with 64K input tokens, output length 1.}
  \label{fig:expert_imbalance}
  \vspace{-10pt}
\end{figure}

Figure~\ref{fig:expert_imbalance} shows that in Qwen3-30B-A3B the max/min token count across experts reaches $16.15\times$ aggregated over all 48 MoE layers (Fig.~\ref{fig:expert_imbalance_aggregated}) and is even more skewed within individual layers (Fig.~\ref{fig:expert_imbalance_layer_13}). This imbalance (i) makes per-expert GEMMs small and irregular, reducing Tensor Core utilization; and (ii) turns the heaviest-loaded GPUs into stragglers that dictate layer completion under EP.

\noindent\emph{\underline{Insight \#2.} Distributed parallel execution degrades compute efficiency through small per-device batches and MoE routing-induced expert-load imbalance.} 

\begin{table*}[t]
\centering
\small
\caption{Per-device, per-layer communication cost of distributed parallel strategies for MoE prefill-only inference.
\textit{Notation follows \S\ref{sec:analysis_system} with $P$ denoting the parallel degree. The communication sizes count bytes received per device and assume ring-style implementations of collective primitives and ignore routing imbalance across experts. The last row, \SysBackend, is our proposed design, detailed in \S\ref{sec:system_design}. \SysBackend's background \texttt{AllGather} traffic is reported in Appendix~\ref{app:comm_accounting}.}}
\label{tab:moe_comm_compare}
\scalebox{0.9}{
\begin{tabular}{p{0.1\linewidth} p{0.1\linewidth} p{0.31\linewidth}  p{0.32\linewidth} p{0.09\linewidth}}
\toprule
\textbf{Attention Parallelism} & \textbf{Expert $\;$ $\;$ $\;$ Parallelism} & \textbf{Collective Operations\newline (ring-style)} & \textbf{Per-Device On-Path Traffic\newline (bytes)} & \textbf{On-Path\newline Volume} \\
\midrule
DP & DP &
None &
$\approx 0$ &
Low \\

DP & TP &
1 $\times$ \texttt{AllGather} + 1 $\times$ \texttt{ReduceScatter} &
$\approx 2 (P-1) \cdot B \cdot S \cdot H \cdot b$  &
High \\

DP & EP &
2 $\times$ \texttt{AllToAll} &
$\approx 2 \cdot k \cdot \frac{P-1}{P} \cdot B \cdot S \cdot H \cdot b$ &
High \\

TP & TP &
2 $\times$ \texttt{AllReduce} &
$\approx 4 \cdot \frac{P-1}{P} \cdot B \cdot S \cdot H \cdot b$  &
Medium \\

TP & EP &
1 $ \times$ \texttt{AllReduce} + 2 $\times$ \texttt{AllToAll} &
$\approx (2+2k) \cdot \frac{P-1}{P} \cdot B \cdot S \cdot H \cdot b$  &
High \\

PP & PP &
2 $\times$ Send/Recv &
$\approx 1 \cdot B \cdot S \cdot H \cdot b$  &
Medium \\

SP & TP &
2 $\times$ \texttt{AllGather} + 1 $\times$ \texttt{ReduceScatter} &
$\approx (2 \cdot N_{kv} \cdot d_{h} + 2 \cdot H) \cdot \frac{P-1}{P} \cdot B \cdot S \cdot b $  &
Medium \\

SP & EP &
1 $\times$ \texttt{AllGather} + 2 $\times$ \texttt{AllToAll} &
$\approx (2 \cdot N_{kv} \cdot d_{h} + 2 \cdot k \cdot H) \cdot \frac{P-1}{P} \cdot B \cdot S \cdot b $ &
High \\

\textbf{DP} & \textbf{\SysBackend{}} &
\textbf{1 $\times$ \texttt{AllGather} (Async)} &
$\bm{\approx 0}$ on path (\texttt{AllGather} in background) &
\textbf{Low} \\

\bottomrule
\end{tabular}
}
\vspace{-10pt}
\end{table*}

\subsection{Communication Overhead}
\label{subsection:overhead_parallel_serving}

Distributed execution introduces two on-path overheads per layer: \emph{collective primitives} (\texttt{AllReduce}, \texttt{AllToAll}) that exchange activations, and \emph{auxiliary kernels} (reshaping, token permutation/alignment) that prepare tensors for them. Both sit on the critical path of every layer.

Table~\ref{tab:moe_comm_compare} summarizes per-device, per-layer communication volume for each attention/expert combination. DP+DP is communication-free but memory-expensive; the strategies feasible for large MoE models (DP+EP, TP+EP) incur traffic proportional to $B \cdot S \cdot H$, amplified under EP by the top-$k$ fan-out. The last row, \SysName's DP+\SysBackend{}, achieves near-zero on-path traffic and is a design preview.

\noindent\emph{\underline{Insight \#3.} Distributed parallel execution introduces frequent on-path collective communication and auxiliary kernels, particularly under expert parallelism where per-layer \texttt{AllToAll} traffic scales with the fan-out $k$.} 

Insights~\#1--\#3 stem from coupling expert placement with synchronous activation routing, which \S\ref{sec:system_design} removes.

\section{\SysName: Design Overview}
\label{sec:system_design}

\S\ref{sec:analysis_workload}--\S\ref{sec:analysis_system} surface three \emph{opportunities} from prefill-only workloads (Observations~\#1--\#3) and three \emph{constraints} in existing MoE serving (Insights~\#1--\#3). \SysName\ distills these into three principles (\S\ref{subsec:principles}), realized by a two-tier architecture (\S\ref{subsec:arch}) whose frontend and backend are co-designed (\S\ref{subsec:codesign}).

\subsection{Design Principles}
\label{subsec:principles}

\SysName\ targets three classes of redundancy on the MoE prefill critical path, mirroring its slogan \emph{Zero Redundancy Overheads}:
\textbf{P1~(computation):} strip decoding-stage machinery and co-locate same-prefix requests so each prefix is executed exactly once (Observations~\#1, \#3);
\textbf{P2~(memory):} exploit the compute window to offload expert weights to CPU DRAM and stream them back asynchronously, share prefix KV via affinity routing, and optionally disable KV storage (Observation~\#2, Insight~\#1);
\textbf{P3~(communication):} gather experts \emph{by weight} rather than routing \emph{by activation}, replacing on-path collectives with background weight transfers overlapped with compute (Insights~\#2--\#3).

\subsection{System Architecture and Workflow}
\label{subsec:arch}

\begin{figure*}[h]
  \centering
  \includegraphics[width=\linewidth]{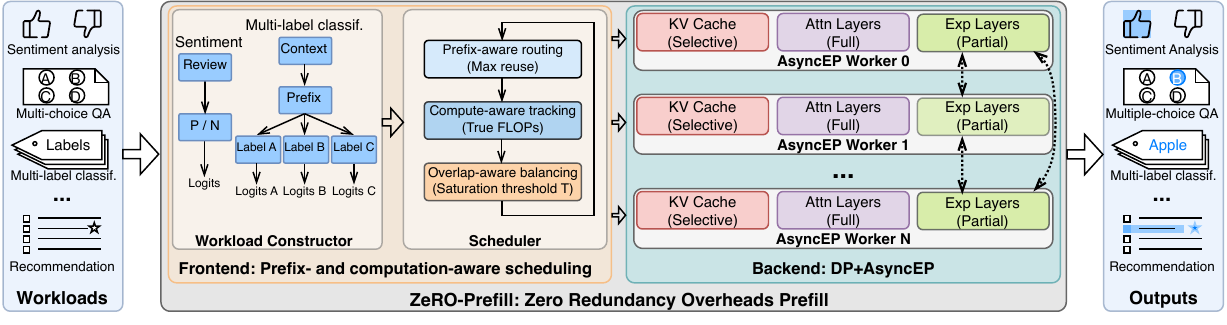}
  \caption{\SysName\ system architecture and end-to-end prefill-only serving workflow.
  \textit{The frontend normalizes incoming tasks into prefill-only form and schedules them into saturation-bounded batches with prefix affinity; the backend executes each batch under data-parallel attention and asynchronous expert streaming, returning logits without entering any decoding loop.}}
  \label{fig:sys_arch}
  \vspace{-5pt}
\end{figure*}

\SysName\ is a two-tier serving system (Figure~\ref{fig:sys_arch}). The frontend (\S\ref{sec:frontend}) first normalizes each request---single-token classification, single-choice selection, or multi-selection decomposed into binary siblings (\S\ref{sec:analysis_workload})---into prefill-only form, then assembles per-GPU batches under three rules: co-locate same-prefix requests to maximize KV reuse, measure load in true FLOPs after prefix-sharing credit, and mark a GPU saturated once its accumulated FLOPs reach the backend-derived \emph{saturation threshold} $T$ (Eq.~\eqref{eq:threshold}). The backend (\S\ref{sec:backend}) runs each batch as pure DP attention with \SysBackend: expert weights for upcoming MoE layers are gathered in the background via D2D \texttt{AllGather} over NVLink (and optionally prefetched from CPU DRAM via PCIe), while the current layer computes. Once the frontend fills a GPU to $T$, the per-layer compute window dominates the slowest ongoing transfer, so no collective appears on the critical path. A KV-cache-free mode further disables KV storage for workloads without prefix reuse.

\subsection{Frontend--Backend Co-Design}
\label{subsec:codesign}

Most LLM serving systems treat the scheduler and the execution engine as independently tunable components connected only by a request queue. \SysName\ departs from this pattern: frontend and backend share three explicit interfaces---the \emph{saturation threshold $T$}, defined by the backend (Eq.~\eqref{eq:threshold}) and enforced by the frontend (\S\ref{subsec:overlap_balancing}); \emph{prefix-KV affinity}, with routing and the KV block table sharing one per-GPU cache (\S\ref{subsec:prefix_routing}); and a \emph{true-FLOPs cost model} that mirrors the backend's prefix and suffix passes (\S\ref{subsec:compute_tracking}).
All three interfaces originate in the backend, so we present it (\S\ref{sec:backend}) before the frontend (\S\ref{sec:frontend}).

\section{Backend: Asynchronous Expert-Parallel Execution}
\label{sec:backend}

The backend of \SysName\ executes the computation scheduled by the frontend, with the goal of keeping every GPU on useful compute rather than waiting on data transfers. We present it as a progression of execution modes, each relaxing one more constraint from \S\ref{sec:analysis_system}.

\subsection{Design Constraints}
\label{subsec:backend_constraints}

\begin{figure}[t]
    \centering
    \includegraphics[width=\linewidth]{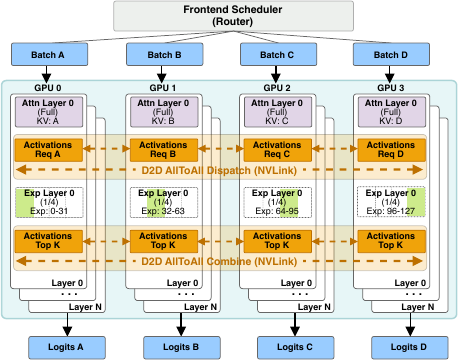}
    \caption{Conventional synchronous DP+EP with four GPUs. Each GPU holds $\frac{1}{N}$ of the experts per layer; two synchronous \texttt{AllToAll}s per MoE layer sit on the critical path.}
    \label{fig:backend_dp_ep}
    \vspace{-10pt}
\end{figure}

Building on Insights~\#1--\#3 (\S\ref{sec:analysis_system}), the backend must jointly satisfy, in descending order of priority: C1~(feasibility) the model's weights, KV cache, and activations fit per-GPU HBM; C2~(performance) no synchronous collective sits on the per-layer critical path; and C3~(balance) no GPU becomes a per-layer straggler under expert-load skew. C1 determines \emph{whether} the model runs, C2 \emph{how fast} it runs, and C3 \emph{how evenly} GPUs utilize compute. Conventional synchronous DP+EP (Figure~\ref{fig:backend_dp_ep}) violates all three: every GPU permanently stores $\frac{1}{N}$ of every layer's experts (C1), two \texttt{AllToAll}s per layer block computation (C2), and the barrier amplifies routing skew into stragglers (C3). \SysName's backend addresses them through a progressive design: \SysBackend via D2D (\S\ref{subsec:async_ep_d2d}) eliminates C2 and C3; \SysBackend with offloading (\S\ref{subsec:async_ep_offload}) additionally relaxes C1 on \emph{weights}; and a KV-cache-free mode (\S\ref{subsec:no_kv_cache}) further relaxes C1 on \emph{KV} for prefix-sparse workloads.

\subsection{\SysBackendName}
\label{subsec:async_ep_d2d}

\SysBackend eliminates activation routing from the critical path by restructuring how expert weights are managed. The key insight is that in prefill-only serving, a GPU does not need to hold experts for all layers simultaneously---it only needs the \emph{complete} set of experts for the layer currently being computed.

\begin{figure*}[t]
    \centering
    \begin{subfigure}[t]{0.49\textwidth}
        \centering
        \includegraphics[width=\linewidth]{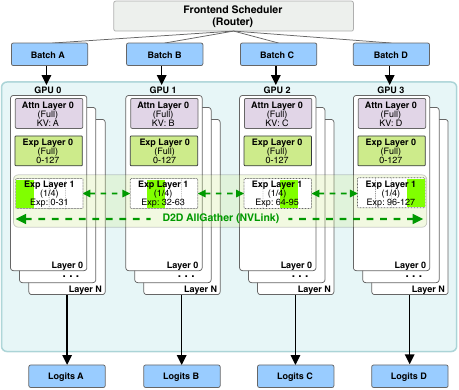}
        \caption{DP+\SysBackend (D2D only)}
        \label{fig:backend_d2d}
    \end{subfigure}
    \hfill
    \begin{subfigure}[t]{0.49\textwidth}
        \centering
        \includegraphics[width=\linewidth]{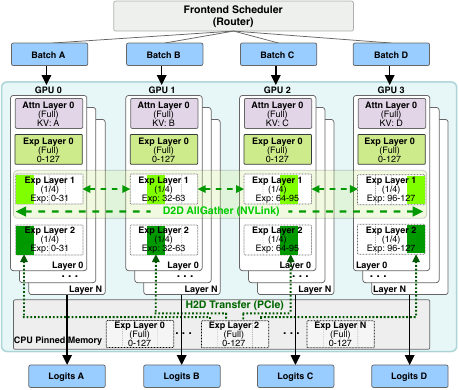}
        \caption{DP+\SysBackend with offloading (H2D+D2D)}
        \label{fig:backend_h2d_d2d}
    \end{subfigure}
    \caption{\SysBackend execution models with four GPUs. (a)~Each GPU replicates all experts for the first layer and gathers subsequent layers' experts via D2D \texttt{AllGather} over NVLink, fully overlapped with current-layer computation. (b)~With offloading, each GPU retains $\frac{1}{N}$ expert shards for only a few upcoming layers; D2D \texttt{AllGather} assembles the next layer while H2D transfers over PCIe simultaneously prefetch further-ahead layers from CPU pinned memory.}
    \label{fig:backend_async_ep}
  \vspace{-10pt}
\end{figure*}

\noindent\textbf{Weight layout and execution.}
As shown in Figure~\ref{fig:backend_d2d}, each GPU holds a full replica of attention weights (identical to DP+EP) and $\frac{1}{N}$ of the expert weights partitioned by expert index, with all GPUs additionally replicating the \emph{complete} expert set for the first MoE layer. After computing attention locally, each GPU evaluates the current MoE layer entirely on-device and simultaneously initiates a background D2D \texttt{AllGather} over NVLink to collect the next layer's complete expert weights from the $\frac{1}{N}$ shards distributed across GPUs. By the time the current layer finishes, the gather has completed and the GPU is ready to execute the next layer locally; the process repeats layer by layer.

\noindent\textbf{Why the overlap works: the saturation threshold $T$.}
For a layer's D2D \texttt{AllGather} to be fully hidden, the layer's compute must last at least as long as the gather. Let $t_\text{EP}$ denote the per-layer time of the slowest EP data transfer (D2D \texttt{AllGather}, or H2D PCIe transfer when offloading is enabled in \S\ref{subsec:async_ep_offload}) and $F_\text{ach}$ the GPU's \emph{achieved} FLOP rate on prefill batches. Each layer must then carry at least $t_\text{EP} \cdot F_\text{ach}$ FLOPs; since all $L$ layers process the same batch, a GPU's per-batch compute fully overlaps the transfers whenever its forward FLOPs exceed the \emph{saturation threshold}
\begin{equation}
\label{eq:threshold}
T = \gamma \cdot L \cdot t_\text{EP} \cdot F_\text{ach},
\end{equation}
where $\gamma \geq 1$ (e.g., $1.2$) absorbs transient jitter. $T$ is a physical quantity calibrated once at startup by a profile run (Appendix~\ref{app:impl_threshold}; values in Appendix~\ref{app:calibrated_T}). Hiding the transfers takes modest load: on 8$\times$H200 with Qwen3-235B-A22B-FP8, a single 8K-token prompt per GPU already hides the D2D transfers (16K with H2D+D2D); below $T$, transfers are only partially exposed, with no batching delay or fallback to synchronous \texttt{AllToAll} (Appendix~\ref{app:envelope_sweep}). Prefill-only workloads naturally exceed $T$ because aggressive batching produces long, compute-bound per-layer forward passes, and the frontend (\S\ref{sec:frontend}) enforces $T$ as the per-GPU admission condition, closing the co-design loop. Two \texttt{AllToAll}s per MoE layer thus disappear from the critical path (C2), and routing imbalance no longer creates per-layer stragglers because all top-$k$ dispatch is local (C3).

\subsection{\SysBackend with Offloading and KV-Cache-Free Execution}
\label{subsec:async_ep_offload}

The D2D-only \SysBackend resolves C2 and C3 but does not fully address C1: each GPU must still permanently store $\frac{1}{N}$ of expert weights for \emph{every} layer, a footprint that for very large MoE models crowds out the HBM available for KV caches and activations. \SysName\ therefore extends \SysBackend with two independent \emph{memory knobs}---hybrid weight offloading on the \emph{weights} dimension and KV-cache-free execution on the \emph{KV} dimension---that shrink the per-GPU HBM footprint and extend deployment to cheaper hardware.

\noindent\textbf{Weights dimension: hybrid offloading.}
As shown in Figure~\ref{fig:backend_h2d_d2d}, each GPU now retains $\frac{1}{N}$ expert weights for only a small number of upcoming layers, a fixed window (e.g., 4 of 94 layers for Qwen3-235B-A22B) reserved at initialization (Appendix~\ref{app:impl_backend}, \ref{app:hbm}); the complete expert weights for all layers live in CPU pinned memory as an offloaded backing store. Execution becomes a \emph{two-channel pipeline}: while the current layer computes, a D2D \texttt{AllGather} over NVLink assembles the immediately-next layer's complete expert set from shards already resident on each GPU, \emph{and} an H2D PCIe prefetch concurrently transfers $\frac{1}{N}$ shards for further-ahead layers from CPU memory, staging them for future \texttt{AllGather}s. The NVLink and PCIe channels run in parallel, so by the time each \texttt{AllGather} starts, the shards it needs are already on-GPU. Because the slowest per-layer transfer now shifts from D2D \texttt{AllGather} to H2D PCIe, the saturation threshold $T$ in Eq.~\eqref{eq:threshold} automatically absorbs this by setting $t_\text{EP}$ to the H2D latency---the frontend still schedules enough compute to cover the slower channel.

\noindent\textbf{KV dimension: KV-cache-free execution.}
\label{subsec:no_kv_cache}
For workloads without significant prefix sharing (e.g., classification over diverse documents), prefix KV caches offer no reuse benefit but still consume substantial HBM. \SysName\ supports a \emph{KV-cache-free} mode that disables KV storage entirely and computes attention on the fly~\cite{flashattention}. The trade-off is explicit---high-prefix-reuse workloads lose the reuse dividend---but for prefix-sparse workloads the KV footprint is pure overhead.

The two knobs are orthogonal and compose: with both enabled, each GPU's HBM holds only attention weights, a few layers of expert shards, and activation workspace---a fraction of static EP's footprint (C1). Freed memory is reinvested in larger KV caches and more aggressive batching, or traded for lower-HBM GPUs that broaden the deployable envelope. Together with \SysBackend (C2+C3), the three components form a progressive, composable design that executes with zero on-path communication, provided the frontend meets the saturation threshold $T$, which is the co-design interface.

\section{Frontend: Prefix-, Compute-, and Overlap-Aware Scheduling}
\label{sec:frontend}

The backend (\S\ref{sec:backend}) hides communication behind compute via \SysBackend transfers. The frontend's role is to make this overlap effective in practice: supply each GPU with \emph{enough} compute to cover the transfer latency, while simultaneously maximizing prefix reuse and avoiding redundant work.

\subsection{Design Constraints for Prefill-Only Scheduling}
\label{subsec:scheduling_constraints}

Combining workload opportunities (\S\ref{sec:analysis_workload}) with backend properties (\S\ref{sec:backend}), the frontend must jointly satisfy: F1~(prefix reuse) co-locate same-prefix requests so the prefix is computed and cached exactly once (Observation~\#3); F2~(compute awareness) account for prefix-sharing credit, since the true cost of $N$ same-prefix requests is one prefix pass plus $N$ suffix passes, not $N$ full ones; and F3~(overlap) keep each GPU's per-batch FLOPs above the saturation threshold $T$ (Eq.~\eqref{eq:threshold}) so \SysBackend{}'s transfers stay hidden on the critical path. F1 decides \emph{where} a request goes, F2 \emph{how} to measure its cost, and F3 \emph{when} a GPU has enough work.

\begin{figure}[t]
    \centering
    \includegraphics[width=\linewidth]{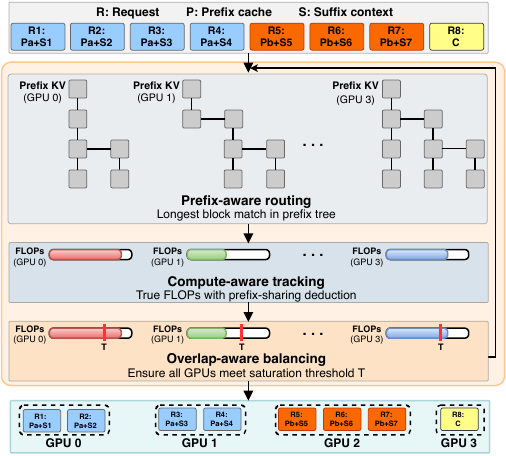}
    \caption{\SysName\ frontend scheduling with four GPUs, realized in three stages:
    (1)~\emph{Prefix-aware routing} picks the GPU with the longest block-level cache match;
    (2)~\emph{Compute-aware tracking} updates each GPU's true-FLOPs load after prefix-sharing credit;
    (3)~\emph{Overlap-aware balancing} marks a GPU saturated once its load reaches the backend-derived threshold $T$.}
    \label{fig:frontend_overview}
  \vspace{-10pt}
\end{figure}

\noindent\textbf{Why existing schedulers fail.}
Existing LLM serving systems optimize mixed prefill--decode workloads with simple load proxies (e.g., vLLM's $\textit{waiting}{\times}4 + \textit{running}$) that work only when per-request cost is dominated by decoding. They violate all three constraints: they ignore prefix sharing and scatter same-prefix requests (F1); even a token-count metric would estimate $10 \cdot (4096{+}S_\text{sfx})$ tokens for 10 same-prefix requests whose true cost is one prefix pass plus 10 short suffix passes, perversely driving the scheduler to disperse them (F2); and they lack a backend-derived overlap threshold, so decoding-calibrated heuristics are misaligned with \SysBackend (F3).

\subsection{Prefix-Aware Routing}
\label{subsec:prefix_routing}

To satisfy F1, \SysName\ routes each request to the GPU whose resident prefix KV cache shares the \emph{longest common prefix} with the request's input. Matching operates at \emph{block granularity} (e.g., $b_{\text{blk}}{=}16$ tokens), aligned with the KV management unit used by PagedAttention~\cite{vllm} and RadixAttention~\cite{sglang}: each block is hashed and the scheduler queries each GPU's block table for the longest contiguous match. This reduces matching cost from $O(P)$ token-level lookups to $O(P/b_{\text{blk}})$ block-level lookups and reuses the block table already maintained by the serving engine.

\noindent\textbf{In-batch vs.\ cross-batch reuse.}
The same longest-match rule covers two time scales without any additional machinery, though the savings differ. \emph{Within a round} (in-batch), sibling requests share a single prefix forward pass: the prefix self-attention is identical across siblings and is computed once, and a single prefix KV copy is stored in HBM. \emph{Across rounds} (cross-batch), each GPU's block table retains prefix KV blocks under LRU eviction; a later request that hits an already-resident prefix reuses its cached KV blocks. Both behaviors emerge from a single routing policy, with no explicit batch-assembly or replication mechanism; the precise FLOPs are captured by $C_\text{pfx}(P_r{-}M_r) + C_\text{sfx}(S_r, P_r)$ in \S\ref{subsec:compute_tracking}.

\subsection{Compute-Aware Tracking}
\label{subsec:compute_tracking}

To satisfy F2, the scheduler maintains a running cost estimate $L_i$ per GPU that reflects actual FLOPs after prefix reuse. When request $r$ with prefix length $P_r$ and suffix length $S_r$ is assigned to GPU~$i$ with $M_r$ prefix tokens already cached, the cost increment is
\begin{equation}
\label{eq:cost_delta}
\Delta_r = C_\text{pfx}(P_r - M_r) + C_\text{sfx}(S_r,\, P_r),
\end{equation}
where $C_\text{sfx}(S, P) = C_\text{FFN}(S) + C_\text{self}(S) + C_\text{cross}(S, P)$ combines linear feed-forward, $O(S^2)$ suffix self-attention, and $O(S{\cdot}P)$ cross-attention of suffix tokens to prefix KV. When the full prefix is cached ($M_r = P_r$), the prefix term vanishes and only the suffix cost remains. After each assignment, $L_i \!\leftarrow\! L_i + \Delta_r$.

This cost model correctly captures prefix-sharing savings: assigning 10 same-prefix requests to a GPU adds one prefix cost plus 10 suffix costs, not 10 full-request costs. The scheduler therefore treats such a GPU as \emph{lightly} loaded---the opposite of a token-based metric---and keeps routing same-prefix requests there until saturation.

\subsection{Overlap-Aware Balancing}
\label{subsec:overlap_balancing}

To satisfy F3, the scheduler enforces the saturation threshold $T$ (Eq.~\eqref{eq:threshold}) derived from the backend's EP communication latency.

At each scheduling round, the scheduler drains a global FIFO queue. For each request, it evaluates all \emph{unsaturated} GPUs ($L_i < T$), selects the one with the longest block-level prefix match (F1), updates its load via the true-cost model (F2), and marks it \emph{saturated} once $L_i \geq T$. The round ends when all GPUs are saturated or the queue is exhausted. A popular prefix that saturates the primary GPU early naturally spills to the next-best-matching GPU through the same logic---no explicit replication policy is needed. Because GPUs are marked saturated one at a time, each GPU's final load lies in $[T,\; T + \Delta_\text{last}]$, so load balance emerges \emph{for free} as a structural byproduct of saturation, not an explicit objective. Pseudocode of one scheduling round is given in Appendix~\ref{app:alg_frontend}.

\section{Evaluation}
\label{sec:evaluation}

\begin{table*}[t]
\centering
\small
\begin{tabular}{llllll}
\toprule
\textbf{Dataset} & \textbf{Workload} & \textbf{Req Context (tokens)} & \textbf{\# Requests} & \textbf{Total (tokens)} & \textbf{Prefix Share} \\
\midrule
MoralStories & Moral reasoning            & $\sim$100--200  & 24K  & $\sim$3.3M   & High \\
MMLU         & Multiple-choice QA         & $\sim$50--500   & 24K  & $\sim$5.8M   & Low \\
BoolQ        & Binary QA                  & $\sim$200--600  & 12K  & $\sim$2.7M   & Low \\
IMDB         & Sentiment classification   & $\sim$300--2K   & 12K  & $\sim$5.5M   & Low \\
QuALITY      & Long-document QA           & $\sim$4K--12K   & 1.2K & $\sim$10.5M  & High \\
ArXiv Class. & Document classification    & $\sim$6K--128K  & 600  & $\sim$10.1M  & Medium \\
\midrule
\textbf{Aggregated} & \textbf{Prefill-only} & \textbf{$\sim$50--128K} & \textbf{73.8K} & \textbf{$\sim$37.9M} & \textbf{Mixed} \\
\bottomrule
\end{tabular}
\caption{Composition of the aggregated prefill-only workload. \textit{Per-source rows describe the contribution of each benchmark; the aggregated workload (last row) is the sole workload used in \S\ref{subsec:eval_e2e}--\S\ref{subsec:eval_ablation}. All requests are reformulated as prefill-only; output length is fixed to 1.}}
\label{table:dataset}
\vspace{-5pt}
\end{table*}

We evaluate \SysName\ along five axes: end-to-end throughput on real-world prefill-only workloads (\S\ref{subsec:eval_e2e}), the contribution of each co-design tier (\S\ref{subsec:eval_ablation}), generalization to synthetic workloads without prefix reuse (\S\ref{subsec:eval_generalization}), memory scalability and per-GPU compute efficiency (\S\ref{subsec:eval_mfu}), and correctness of the prefill-only reformulation (\S\ref{subsec:eval_accuracy}).

\subsection{Evaluation Setup}
\label{subsec:eval_setup}

\noindent\textbf{Datasets.}
To reflect the heterogeneity of production prefill-only traffic, we construct a single \emph{aggregated workload} by mixing requests sampled from six public benchmarks (Table~\ref{table:dataset}): short-context classification and QA (MoralStories~\cite{moralStories}, MMLU~\cite{mmlu}, BoolQ~\cite{boolq}), medium-length document analysis (IMDB~\cite{imdb}, QuALITY~\cite{quality}), and long-context document classification (ArXiv Classification~\cite{arxiv_class}, up to 128K tokens). Together they span the full prefix-share spectrum---from high (shared system prompts) through medium to low (independent documents)---and yield a workload of 73.8K requests and $\sim$37.9M tokens, mirroring a production endpoint that interleaves classification, QA, and long-document requests.

\noindent\textbf{Hardware, models, and baselines.}
We evaluate \SysName\ on three datacenter GPU platforms---$8{\times}$A100
(80GB, BF16), $8{\times}$H100 (80GB, BF16 and FP8), and $8{\times}$H200
(141GB, FP8)---using \emph{Qwen3-235B-A22B}~\cite{qwen3} (128 experts, top-8 routing,
$\sim$22B activated parameters). 
We compare against two groups of baselines, swept at parallel degrees 1, 2, 4, and 8.
\textit{Group~(i)}: the five distributed strategies natively supported
by vLLM for large-MoE inference---DP+EP, DP+TP, TP+EP, TP+TP, and
PP+PP (Table~\ref{tab:moe_comm_compare}).
\textit{Group~(ii)}: four PrefillOnly-augmented configurations.
PrefillOnly~\cite{prefillonly} targets models that fit on a single GPU; its
hybrid-prefilling and suffix-KV-cache-discarding techniques are
engineered to enable such deployment by reducing per-GPU memory.
Since Qwen3-235B-A22B exceeds the HBM of every GPU tested, these techniques have a negligible impact.
We therefore port its scheduler, shortest-prefill-first, with continuous prefill-time estimation into
vLLM~v0.11.0 and pair it with two distributed backends, DP+EP and
PP+PP (the strongest Group~(i) baseline), each at two batch sizes:
\texttt{max-num-seqs=1} (PrefillOnly's paper default setting) and a
value matched to \SysName.
For every hardware--precision cell we report \SysName's improvement
over the best baseline per parallel degree.

\noindent\textbf{Implementation and metrics.}
\SysName\ is implemented on top of vLLM (details in Appendix~\ref{app:implementation}) and enabled via a single flag. We report (i) throughput in tokens/s (total prefilled tokens divided by end-to-end batch completion time); (ii) MFU as achieved FLOPs over the GPU's peak FLOPs at the deployed precision; and (iii) task accuracy. All runs fix the output length to 1, isolating prefill execution.

\subsection{End-to-End Throughput on Real-World Workloads}
\label{subsec:eval_e2e}

\begin{figure*}[t]
    \centering
    \includegraphics[width=\linewidth]{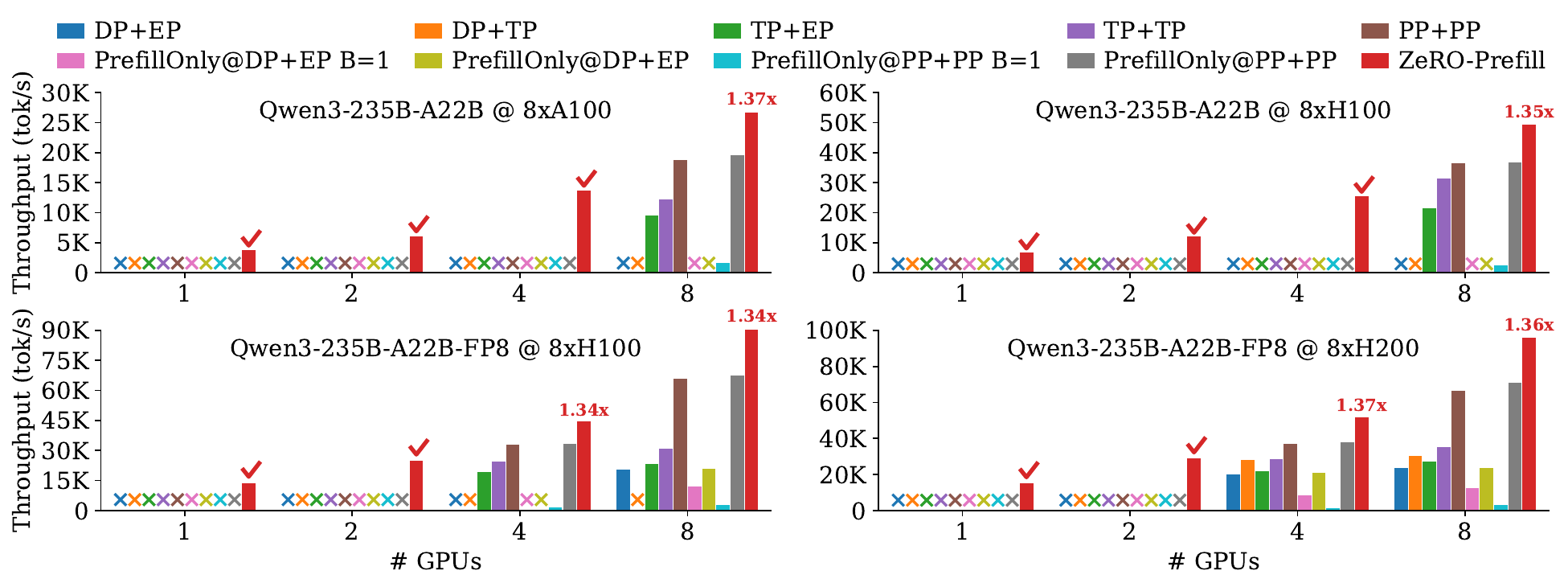}
    \caption{End-to-end throughput on the aggregated real-world prefill-only workload. \textit{Each panel reports one hardware--precision combination; bars compare the five vLLM distributed baselines and the four PrefillOnly configurations (\S\ref{subsec:eval_setup}) with \SysName\ at parallel degrees 1/2/4/8. Input lengths span 50--128K tokens; output length is fixed to 1.}}
    \label{fig:throughput_comparison_real-world_all}
\end{figure*}

Figure~\ref{fig:throughput_comparison_real-world_all} reports throughput on the aggregated real-world workload across all four hardware--precision combinations, sweeping the parallel degree from 1 to 8 GPUs.

\noindent\textbf{Uniform dominance across precisions.}
\SysName\ achieves the highest throughput in every hardware--precision--parallel-degree cell, delivering \textit{1.37$\times$ on 8$\times$A100 (BF16), 1.35$\times$ on 8$\times$H100 (BF16), 1.34$\times$ on 8$\times$H100 (FP8), and 1.36$\times$ on 8$\times$H200 (FP8)} over the strongest baseline at 8 GPUs.
\SysName's $\sim$1.35$\times$ relative gain is stable even as FP8 roughly doubles absolute throughput, consistent with the analytical prediction in \S\ref{sec:analysis_system} that on-path \texttt{AllToAll} traffic scales with $B{\cdot}S{\cdot}H$ independent of element size.

\noindent\textbf{Near-linear scaling.}
Baselines exhibit the sub-linear scaling predicted by Table~\ref{tab:moe_comm_compare}: adding GPUs amortizes the weight footprint but injects more per-layer collective traffic, and several baselines plateau or regress past 4 GPUs. \SysName\ scales near-linearly because its only cross-GPU traffic, \SysBackend{}'s D2D \texttt{AllGather}, is background-overlapped with compute, decoupling throughput from collective-scaling limits.

\begin{figure*}[t]
    \centering
    \includegraphics[width=\linewidth]{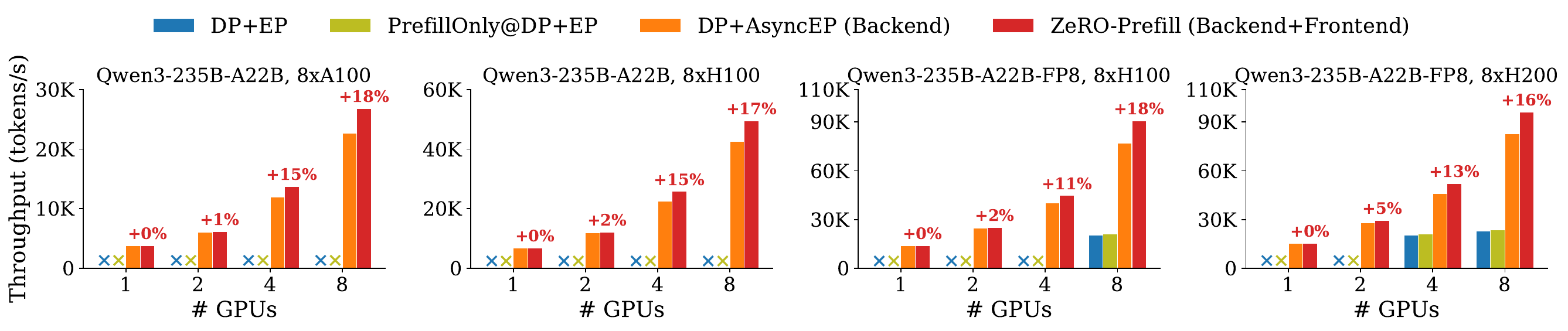}
    \caption{Contribution of \SysName's two design tiers on the real-world workload. \textit{DP+\SysBackend\ applies the backend of \S\ref{sec:backend} under vLLM's default scheduler; \SysName\ additionally applies the frontend of \S\ref{sec:frontend}. Annotations report the throughput gain of adding the frontend over the backend-only configuration at each parallel degree.}}
    \label{fig:throughput_ablation}
  \vspace{-10pt}
\end{figure*}

\subsection{Necessity of Frontend--Backend Co-Design}
\label{subsec:eval_ablation}

Because DP+\SysBackend\ (backend only) already resolves the on-path-collective bottleneck (\S\ref{sec:backend}), a natural question is whether the frontend of \S\ref{sec:frontend} contributes enough to justify its complexity. Fig.~\ref{fig:throughput_ablation} decomposes the gain into two tiers: DP+EP (synchronous MoE), DP+\SysBackend\ under vLLM's default scheduler, and \SysName\ (backend + frontend co-design).

Fig.~\ref{fig:throughput_ablation} annotates each bar with the throughput gain of adding the frontend (i.e., \SysName\ vs.\ DP+\SysBackend\ under vLLM's default scheduler). At 8 GPUs, the frontend improves throughput by \textit{+18\% (A100, BF16), +17\% (H100, BF16), +18\% (H100, FP8), and +16\% (H200, FP8)}. The contribution \emph{grows with parallel degree}: at low GPU counts, random dispatch still achieves reasonable prefix hits, but adding GPUs scatters same-prefix requests and dilutes reuse, whereas \SysName's longest-block-match routing sustains high reuse at any scale.

\subsection{Generalization Across Context Regimes}
\label{subsec:eval_generalization}

To isolate the effect of context length \emph{independent} of prefix reuse, we sweep synthetic workloads with uniformly random prompts (no prefix sharing) across four context regimes: short ($256{\times}40960$), medium ($4\text{K}{\times}2560$), long ($32\text{K}{\times}320$), and ultra-long ($128\text{K}{\times}80$), where $S{\times}N$ denotes sequence length $\times$ number of requests (10M tokens per regime).

\begin{figure*}[t]
    \centering
    \includegraphics[width=\linewidth]{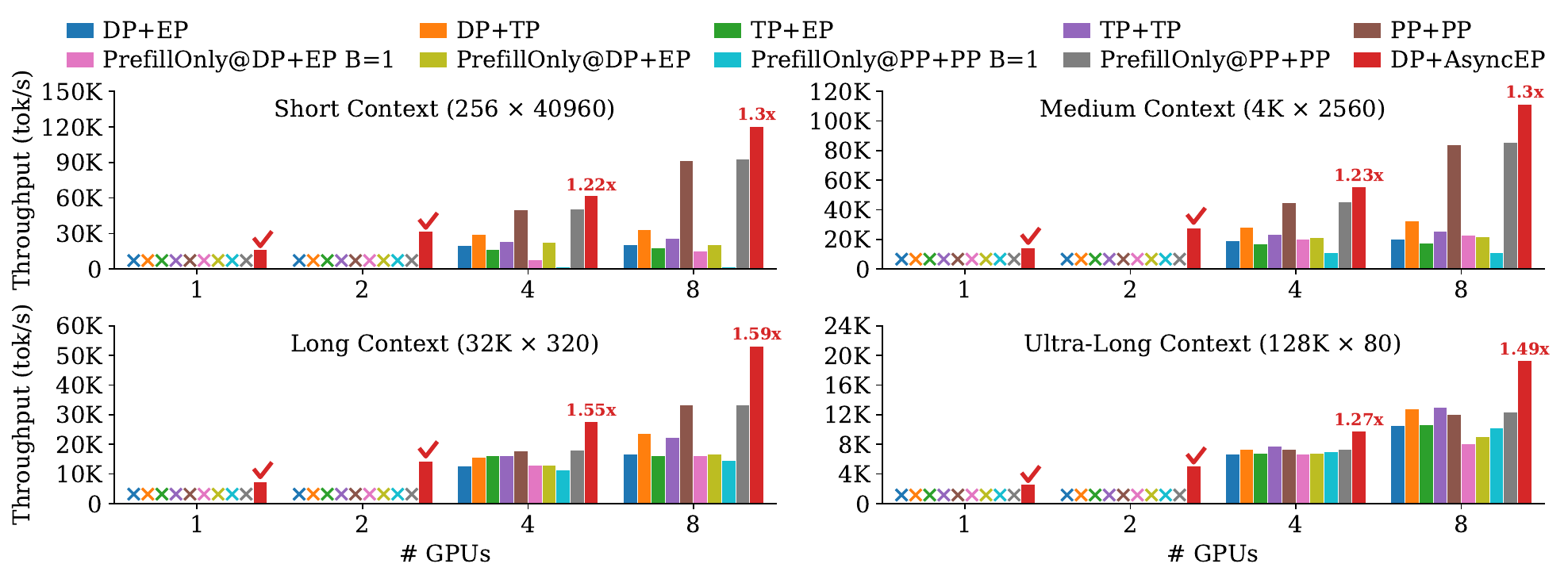}
    \caption{Throughput under synthetic workloads with no prefix reuse, across four context regimes on 1--8 H100 GPUs (FP8). \textit{All panels share the same $S{\times}N$. DP+\SysBackend\ is the backend alone (no prefix-aware routing), isolating the backend contribution.}}
    \label{fig:throughput_comparison_h100_random}
\end{figure*}

\SysName's backend alone (DP+\SysBackend) improves over the best baseline by \textit{1.30$\times$ (short), 1.30$\times$ (medium), 1.59$\times$ (long), and 1.49$\times$ (ultra-long)} at 8 GPUs. Even without prefix reuse, \SysBackend\ delivers $\geq$1.30$\times$, confirming that P3 (zero redundant communication) produces gains independent of P1. The improvement is \emph{larger} at long and ultra-long contexts because the baselines degrade on two fronts: on-path \texttt{AllToAll} volume scales linearly with $B{\cdot}S$, making EP-based strategies proportionally more costly; and under the baselines' default chunked prefill, later chunks of a long prompt attend to longer prefixes and cost more, so PP+PP pipelines micro-batches of unequal cost and exposes bubbles. \SysName\ is immune to both: \SysBackend{}'s weight transfers are fully hidden behind compute regardless of sequence length, and pure DP attention eliminates pipeline partitioning. The same trend holds on GLM-4.5-FP8, a second MoE family: DP+\SysBackend\ improves over the best baseline by 1.18$\times$ (short) to 1.57$\times$ (long) at 8 GPUs (Appendix~\ref{app:new_model}).

\begin{figure*}[t]
  \centering
  \includegraphics[width=\linewidth]{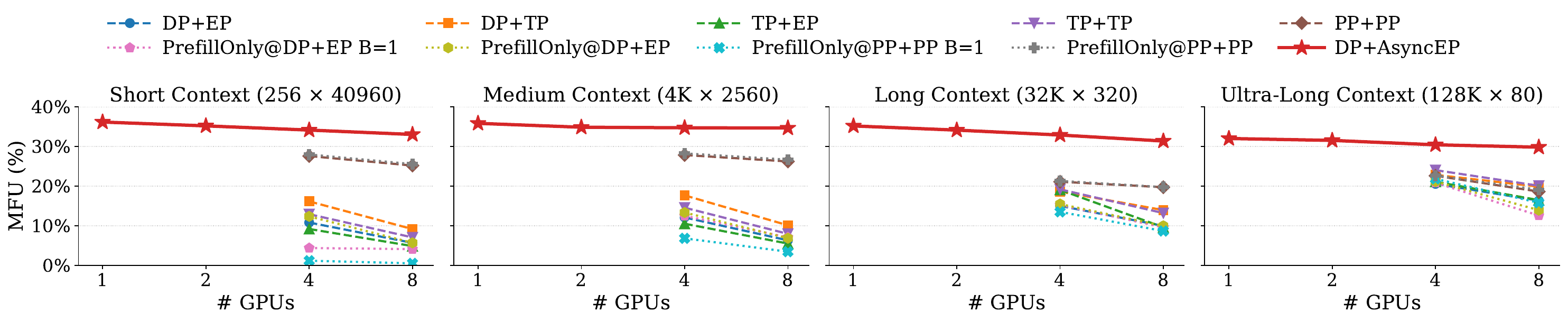}
  \caption{MFU on Qwen3-235B-A22B (H100, FP8) under synthetic no-prefix-reuse workloads, across four context regimes and parallel degrees 1/2/4/8. \textit{Baselines have no points at 1--2 GPUs because the model does not fit; only DP+\SysBackend\ runs there.}}
  \label{fig:MFU_comparison_h100_random}
\end{figure*}

\subsection{Memory Scalability and Compute Efficiency}
\label{subsec:eval_mfu}

Distributed MoE serving usually sacrifices efficiency on two axes at once: HBM capacity (235B-class models require multi-GPU deployment just to hold the weights) and per-GPU MFU (smaller per-device batches fall below the GEMM-saturation threshold, and on-path collectives steal cycles from useful compute; Insight~\#2). \SysName's P2 (hybrid offloading + KV-cache-free execution) and P3 (\SysBackend) attack the two bottlenecks jointly. We reuse the synthetic sweep of \S\ref{subsec:eval_generalization} (8$\times$H100 FP8, Qwen3-235B-A22B, four context regimes, parallel degrees 1/2/4/8), measuring MFU as achieved FLOPs normalized by the 1979~TFLOPS FP8 peak reported in the H100 datasheet~\cite{nvidia_h100}.

\noindent\textbf{Memory scalability at no MFU cost.}
Qwen3-235B-A22B occupies 235~GB in FP8 (470~GB in BF16), which exceeds the aggregate 160~GB HBM of two H100s even before activations. Every baseline in Table~\ref{tab:moe_comm_compare} therefore requires $\geq$4 GPUs simply to hold the weights---the missing baseline points at 1--2 GPUs in Fig.~\ref{fig:MFU_comparison_h100_random} make this gap concrete. Under P2, \SysName\ retains only attention weights plus a sliding window of upcoming expert shards on-GPU while streaming the rest from CPU pinned memory over PCIe, combined with KV-cache-free execution for prefill-only workloads; the per-GPU HBM footprint then fits within 80~GB, pushing the deployable envelope from ``$\geq$4 GPUs'' down to a single commodity 80~GB GPU. Crucially, this does not come at a compute-efficiency cost: on 1--2 GPUs \SysName\ sustains \textit{32.0--36.2\% MFU}, comparable to its 29.8--34.7\% MFU at 8 GPUs, because the large per-device token batch in the memory-constrained regime fully overlaps H2D expert transfers behind compute. Appendix~\ref{app:hbm} breaks down per-GPU memory by mode: with a 128K context on 8$\times$H200, hybrid mode leaves 1.5$\times$ (FP8) and 2.0$\times$ (BF16) the KV capacity of DP+EP.

\noindent\textbf{Dominance in the multi-GPU regime.}
From 4 to 8 GPUs, every baseline degrades monotonically in every context regime (e.g., DP+EP drops 1.90$\times$ at short context), confirming the two predicted mechanisms: sub-saturation GEMMs and on-path collective overhead. DP+\SysBackend\ is the top bar in \emph{every} (context, parallel degree) cell at \textit{29.8--36.2\% MFU}, beating the strongest baseline by \textit{1.30$\times$ (short), 1.30$\times$ (medium), 1.59$\times$ (long), 1.49$\times$ (ultra-long)} at 8 GPUs. The feasible hardware envelope therefore widens from ``$\geq$4 GPUs'' to ``1--8 GPUs'' (a 4$\times$ broader range), and within that entire envelope per-GPU MFU varies by only 1.03--1.09$\times$---validating Insight~\#2 and the zero-redundancy design of P2+P3.

\subsection{Accuracy of Prefill-Only Reformulation}
\label{subsec:eval_accuracy}

Correctness decomposes into two orthogonal claims mirroring the two-tier design of \S\ref{subsec:codesign}. \emph{Tier~1 (execution fidelity):} \SysBackend\ only changes \emph{when} expert weights arrive, hybrid offloading only relocates them between HBM and CPU memory, and KV-cache-free execution recomputes attention under the same formulation; none of these modifies GEMM math, attention kernels, or quantization, so outputs match vLLM up to FP non-associativity by construction; on a 25K-sample run, end-task accuracy matches non-offloaded vLLM within measurement noise (Appendix~\ref{app:accuracy_prod_harness}). \emph{Tier~2 (reformulation fidelity):} on a nine-task production classification harness with two open-source models, prefill-only logit scoring lies within \textit{$\pm$3.6\,pp} of autoregressive decoding on 7 of 9 tasks (Table~\ref{tab:accuracy_prod_harness}); a same-model three-way ablation (Table~\ref{tab:accuracy_imdb_ablation}) isolates the decoding-mode contribution to 2.3\,pp. Together these validate end-to-end accuracy of prefill-as-a-service under \SysName.

\vspace{-9pt}
\section{Related Work}
\label{sec:related}

\SysName\ intersects four lines of prior work; the structural difference throughout is that \SysName\ gathers experts \emph{by weight} rather than routing activations, co-designed with a physically derived saturation invariant.

\noindent\textbf{LLM inference engines and prefill optimizations.}
Production engines~\cite{vllm, sglang, trtllm, su2026morphserve, deepspeedinference, orca} and techniques such as continuous batching~\cite{orca}, chunked prefill~\cite{sarathi}, speculative decoding~\cite{speculative}, and prefill/decode disaggregation~\cite{splitwise,distserve,mooncake} all presume prefill is amortized over many decoding steps. PrefillOnly~\cite{prefillonly} pioneered prefill-only serving with KV-cache management and job-level scheduling (\S\ref{sec:intro}); \SysName\ instead tackles the execution-level overheads of distributed serving for large MoE models~(\S\ref{sec:analysis_system}).

\noindent\textbf{Mixture-of-experts serving.}
MoE scales capacity via sparse activation~\cite{switchtransformer,lepikhin2020gshard_EP,glam,stmoe} and is adopted by most recent open-weight models~\cite{mixtral,deepseekv3,qwen3,gptoss,glm45}. System-side MoE work~\cite{fastmoe, tutel, deepspeedmoe, megablocks, fastermoe, lina, moelightning, lancet, exflow, pregatedmoe, epsmoe, elastic_moe, elasticmoe} optimizes traffic, placement, or kernels \emph{within} the synchronous activation-routed EP model, keeping two per-layer \texttt{AllToAll}s on the critical path. \SysName\ inverts this model: experts are gathered by weight, removing \texttt{AllToAll} and dissolving routing-imbalance stragglers (\S\ref{sec:backend}).

\noindent\textbf{Prefix caching and attention reuse.}
PagedAttention~\cite{vllm}, RadixAttention~\cite{sglang}, and related caching systems~\cite{lmcache, mooncake, cachedattention, chunkattention, cascadeinference, hydragen} treat prefix reuse as a \emph{passive cache effect}. \SysName's frontend turns it into an \emph{active scheduling decision} via longest-block-match routing, enabled by \SysBackend\ freeing the HBM needed for large batches that make prefix co-location meaningful. BlendServe~\cite{blendserve} overlaps \emph{requests} with complementary resource profiles atop a synchronous engine; \SysName\ overlaps \emph{layers} via async weight gathering, and the two compose.

\noindent\textbf{Expert and weight offloading.}
Dense offloading systems~\cite{flexgen, zeroinference, yu2026faascale, deepspeedzero, sti} and MoE extensions~\cite{pregatedmoe, swapmoe, fiddler, moeinfinity} either transfer on demand (on the critical path) or rely on popularity heuristics that degrade under large-batch routing skew. \SysName\ exploits the \emph{deterministic layer order of prefill} and sizes the compute window via $T$ (Eq.~\eqref{eq:threshold}), making offloaded experts appear always-resident by construction.

\section{Discussion}
\label{sec:discussion}
\vspace{-3pt}

\noindent\textbf{Applicability and limitations.}
\SysName\ targets throughput-oriented, batch-driven prefill-only serving on MoE models that exceed single-GPU HBM. It is \emph{not} designed for latency-critical interactive serving, arrivals too bursty to sustain the saturation threshold, or dense models with no expert stack.
Three limitations merit note: (i)~on low-bandwidth interconnects $t_\text{EP}$ and thus $T$ grow, potentially re-exposing some transfers on the critical path; (ii)~$T$ is calibrated once at startup---severe workload drift can briefly expose part of the transfers on the critical path, though prefill-only batches refill quickly; and (iii)~KV-cache-free mode and prefix-aware routing are workload-dependent knobs that contribute nothing on truly random traffic but do not regress (\S\ref{subsec:eval_generalization}).

\noindent\textbf{Broader implications.}
The core insight---decoupling expert placement from activation routing---applies whenever the per-layer compute window covers transfer latency, including long-context reasoning and speculative-decode verification. The same principle composes with attention--expert disaggregation~\cite{megascaleinfer} (Appendix~\ref{app:discussion}) and benefits emerging prefill-only decision models such as Jev~\cite{jev} (Appendix~\ref{app:jev}).

\section{Conclusion}
\label{sec:conclusion}
\vspace{-3pt}

Prefill-only workloads expose three structural redundancies in distributed MoE serving, all rooted in coupling expert placement with synchronous activation routing. \SysName\ inverts this coupling: \SysBackend\ replaces per-layer \texttt{AllToAll} with background weight \texttt{AllGather} fully overlapped with compute, while a co-designed frontend enforces a saturation threshold to keep transfers hidden. On Qwen3-235B-A22B, \SysName\ delivers $1.34$--$1.37\times$ throughput over the strongest baseline and sustains $29.8$--$36.2\%$ MFU across 1--8 GPUs. The broader lesson: expert weights are best treated not as static parameters but as schedulable resources driven by the execution schedule.

\section*{Acknowledgments}
\label{sec:acknowledgment} 

We thank the anonymous reviewers and our shepherd for their valuable feedback. This research was supported in part by the National Science Foundation (NSF) grants CNS-2322860, OAC-2411009, and OAC-2403313.
We also acknowledge support from NSF CloudBank and the Commonwealth Cyber Initiative (CCI).

\bibliographystyle{plain}
\bibliography{reference}

\newpage
\appendix
\section*{Appendix}

\section{System Implementation}
\label{app:implementation}

We implement \SysName on top of vLLM~\cite{vllm} (version v0.11.0), reusing its request ingestion, tokenizer, and model execution paths. The additions span a backend weight streaming engine (\S\ref{sec:backend}; Appendix~\ref{app:impl_backend}), a frontend DP router (\S\ref{sec:frontend}; Appendix~\ref{app:impl_frontend}), and the calibration of $T$ that links them (Appendix~\ref{app:impl_threshold}), and require no changes to model definitions, kernels, or the tokenizer pipeline.

\subsection{Backend: Weight Streaming Engine}
\label{app:impl_backend}

The backend presents three execution modes---one-GPU H2D streaming, multi-GPU D2D \texttt{AllGather} over NVLink (\S\ref{subsec:async_ep_d2d}), and concurrent H2D prefetch + D2D \texttt{AllGather} (hybrid mode, \S\ref{subsec:async_ep_offload})---carried by two helpers on dedicated CUDA streams. The \emph{MoE gatherer} issues the \texttt{AllGather} for layer $i{+}K$ when layer $i$ completes and synchronizes it with a single event wait at that layer's entry; if the gather is still in flight, the compute stream simply waits on the event, with no fallback to synchronous \texttt{AllToAll}. The \emph{H2D offloader} pins the CPU backing store and prefetches each layer's local weights one layer further ahead than the gatherer. Both helpers share one ring of $S$ GPU-resident slots, with layer~$i$ in slot $i \bmod S$: while layer~$i$ computes, the gatherer fills layer~$i{+}K$ and, in hybrid mode, the offloader layer~$i{+}K{+}1$. Hence $S \geq K{+}1$ ($K{+}2$ in hybrid mode), and $S$ must also align with $L$ so that slots reused across consecutive forwards are not overwritten early; for Qwen3-235B-A22B with $K{=}1$, $S$ can be as low as 2 (D2D-only) or 4 (hybrid). The ring is allocated once at model load, before vLLM sizes the KV cache, and never grows; the remaining HBM goes to the KV cache and activations. In hybrid mode the two pipelines saturate PCIe and NVLink concurrently, and the model runner sees only the slower channel (captured by $t_\text{EP}$ in Eq.~\eqref{eq:threshold}). KV-cache-free execution is an orthogonal launch flag that skips the KV allocator entirely.

\subsection{Frontend: Prefix-Aware DP Router}
\label{app:impl_frontend}
\label{app:alg_frontend}

The frontend replaces vLLM's default weighted-count DP load balancer with an event-driven state machine that tracks, per engine~$i$, (a)~a \emph{committed} table $\mathcal{C}_i$ of block hashes already materialized on the GPU, updated from the engine's block-cache lifecycle events (captures cross-batch reuse); (b)~a \emph{pending} table $\mathcal{P}_i$ of hashes promised by requests routed but not yet executed, updated speculatively at routing time (captures in-batch reuse); and (c)~a FLOPs-denominated work counter $L_i$ computed from a single per-token constant $f_\text{tok}$ derived analytically from the model config. Because both enqueue and decay use the same $f_\text{tok}$, $L_i$ is drift-free even though the engine reports progress in tokens.

For each request, the router applies the constraints of \S\ref{subsec:scheduling_constraints} in one pass: among engines below saturation ($L_i < T$, F3), it picks the one with the longest $\mathcal{C}_i \cup \mathcal{P}_i$ prefix match (F1), breaking ties and no-match cases by the smallest $L_i$; it then adds the request's uncached block hashes to $\mathcal{P}_{i^*}$ and charges $\Delta_r = u_r \cdot f_\text{tok}$ to $L_{i^*}$ (F2), where $u_r$ is the number of uncached prompt tokens. This keeps the linear part of Eq.~\eqref{eq:cost_delta} and omits attention-score terms, which only underestimates load, so a GPU marked saturated still executes at least $T$ FLOPs. The router never holds requests back: under low load a request runs on its engine's next step (Appendix~\ref{app:envelope_sweep}), and if every engine is saturated it still goes to the best-matching engine, which queues it within its per-step token budget (e.g., 128K tokens). Cross-process hash agreement---so $\mathcal{C}_i$ updates reconcile with router-computed hashes---is secured by reusing the engine's own block-hasher and pinning the Python hash seed identically in every subprocess. Separating $\mathcal{C}_i$ from $\mathcal{P}_i$ is what makes in-batch reuse correct under asynchronous routing: a burst of same-prefix requests seeds $\mathcal{P}_i$ on the first arrival and converges to the same engine, and entries are promoted $\mathcal{P}_i \to \mathcal{C}_i$ on the next block-stored lifecycle event.

Algorithm~\ref{alg:frontend} summarizes one scheduling round of \SysName's frontend (\S\ref{sec:frontend}), integrating prefix-aware routing (F1), compute-aware tracking (F2), and overlap-aware saturation (F3) in a single per-request assignment pass.

\begin{algorithm}[!ht]
\small
\DontPrintSemicolon
\SetKwInOut{Input}{Input}
\SetKwInOut{Maintain}{State}

\Input{Global request queue $\mathcal{Q}$; $N$ GPU workers; saturation threshold $T$}
\Maintain{Per-GPU block table $\mathcal{K}_i$}

\BlankLine
\tcp{\textbf{Step 1: Saturation-based assignment}}
$L_i \leftarrow 0$,\; $\mathcal{A} \leftarrow \{1, \ldots, N\}$ \tcp*{Active GPU set}
\ForEach{request $r \in \mathcal{Q}$ in arrival order}{
    \lIf{$\mathcal{A} = \emptyset$}{\textbf{break}}
    $m_i \leftarrow \text{BlockMatch}(\mathcal{K}_i,\; r)$ for each $i \in \mathcal{A}$\;
    $i^* \leftarrow \arg\max_{i \in \mathcal{A}}\, m_i$;\; break ties by $\arg\min_{i \in \mathcal{A}} L_i$\;
    $\Delta \leftarrow C_\text{pfx}(|p_r| - m_{i^*}) + C_\text{sfx}(|s_r|,\, |p_r|)$\;
    Assign $r$ to GPU $i^*$;\; $L_{i^*} \leftarrow L_{i^*} + \Delta$\;
    \lIf{$L_{i^*} \geq T$}{$\mathcal{A} \leftarrow \mathcal{A} \setminus \{i^*\}$}
}

\BlankLine
\tcp{\textbf{Step 2: Execute \& update}}
Each GPU $i$ executes its assigned batch\;
Update $\mathcal{K}_i$ with newly computed prefix KV caches\;

\caption{Frontend scheduling: one round}
\label{alg:frontend}
\end{algorithm}

\subsection{FLOPs-Native Saturation Threshold Calibration}
\label{app:impl_threshold}

The threshold $T$ (Eq.~\eqref{eq:threshold}) is derived from physical measurements taken during the backend's profile run, not from manual tuning, and is expressed in FLOPs---the same unit as $L_i$. The rationale is that compute \emph{time}---the quantity that must dominate \SysBackend transfers---is determined by FLOPs rather than token count; tokens are only a proxy whose accuracy varies with MoE active-parameter count and layer mix.

\noindent\textbf{Per-layer instrumentation.}
During the profile run the model runner arms CUDA-event timing on the existing per-layer entry/exit hooks. After a warm-up forward at the engine's maximum per-step batch size $n_\text{ref}$, which absorbs one-off costs such as kernel JIT, a second forward records $t_c$, the median wall-clock of the first $K$ layers, whose weights are prefetched and thus measure pure compute, and $t_e$, the median wall-clock of the remaining layers, each taking $\max(\text{compute}, \text{transfer})$, so $t_e \geq t_c$ up to measurement noise. The profile pass skips attention, so $t_c$ covers only the linear compute counted by $f_\text{tok}$ (Appendix~\ref{app:impl_frontend}); together with the envelope $t_e$ and the margin $\gamma$ below, this makes $T$ conservative, erring toward hiding the weight transfers (Appendix~\ref{app:calibrated_T}). The hooks are gated by a boolean, so steady-state serving pays nothing for the instrumentation.

\noindent\textbf{Threshold formula.}
Pairing $(t_c, t_e)$ with the analytical FLOPs estimate $C_\text{dummy} = f_\text{tok} \cdot n_\text{ref}$, the frontend computes
\begin{equation}
  T \;=\; \gamma \cdot \frac{t_e}{t_c} \cdot C_\text{dummy} \quad \text{[FLOPs]},
  \label{eq:impl_threshold_flops}
\end{equation}
where $\gamma$ (default $1.2$) adds a safety margin. Eq.~\eqref{eq:impl_threshold_flops} is the calibrated form of Eq.~\eqref{eq:threshold}, which remains the normative definition: $C_\text{dummy}/(L \cdot t_c)$ estimates the achieved rate $F_\text{ach}$, and the envelope $t_e \geq t_\text{EP}$ conservatively bounds the transfer time, so substituting both into Eq.~\eqref{eq:threshold} yields Eq.~\eqref{eq:impl_threshold_flops}. When transfers are fully hidden, Eq.~\eqref{eq:impl_threshold_flops} reduces to about $\gamma \cdot C_\text{dummy}$, i.e., the reference batch alone already covers the transfers. Figure~\ref{fig:calibration_accuracy} validates the calibration on H200 in FP8 and BF16.

\noindent\textbf{Delivery and fallback.}
The payload $(t_c, t_e, C_\text{dummy})$ is piggybacked onto the first post-calibration engine output; the router installs $T$ once every engine has reported, using the most conservative (largest) $t_e/t_c$ across engines so that $T$ covers the slowest engine's transfers. Until calibration arrives, the router uses the conservative fallback $T = \gamma \cdot f_\text{tok} \cdot n_\text{ref}$, equivalent to Eq.~\eqref{eq:impl_threshold_flops} with $t_e/t_c = 1$.

\section{Operating Envelope of \SysBackend}
\label{app:envelope}

This appendix characterizes when \SysBackend's transfers stay hidden: Appendix~\ref{app:calibrated_T} compares the calibrated threshold with the measured overlap, and Appendix~\ref{app:envelope_sweep} shows how compute efficiency behaves around it. Unless noted, numbers here and in Appendix~\ref{app:costs} refer to Qwen3-235B-A22B ($H{=}4096$, $h{=}1536$, $E{=}128$, $k{=}8$, $L{=}94$) with the default slot ring ($S{=}5$, $K{=}3$).

\subsection{Calibrated Thresholds vs.\ Measured Knees}
\label{app:calibrated_T}

\begin{figure}[!t]
  \centering
  \includegraphics[width=\linewidth]{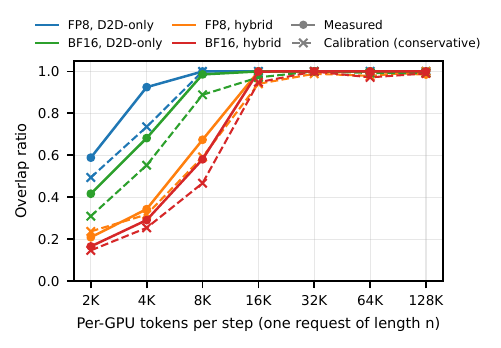}
  \caption{Overlap ratio (share of per-layer time not stalled on weight transfers) versus per-GPU tokens per step for Qwen3-235B-A22B on 8$\times$H200, measured while serving (solid) and estimated by the profile-run calibration (dashed), from which $T$ is computed. The calibration underestimates the overlap at nearly every load, which only raises $T$: the frontend then packs more tokens per step than needed to hide the transfers, leaving headroom at run time.}
  \label{fig:calibration_accuracy}
  \end{figure}

We sweep the per-GPU load per step from 2K to 128K tokens on 8$\times$H200 in FP8 and BF16, each with D2D-only \SysBackend\ and with hybrid offloading. Each step runs one request of length $n$ in KV-cache-free mode (\S\ref{subsec:no_kv_cache}), so the per-step load equals the context length. We time real forward passes and record the overlap ratio, the share of per-layer time in which the compute stream is not stalled on weight transfers; the knee is the smallest load at which it reaches 0.99.

Transfers are exposed only at small loads (Figure~\ref{fig:calibration_accuracy}, solid). Under FP8 D2D-only, a layer's \texttt{AllGather} takes 6.2~ms, so the overlap ratio is 0.59 at 2K and 0.92 at 4K, and the knee lies at 8K. BF16 doubles gathered bytes (12.2~ms per layer) but slows compute by only 1.2--1.6$\times$, moving the D2D-only knee to 16K (0.986 at 8K). Hybrid offloading adds a PCIe transfer per layer (13.4~ms in FP8, 26.8~ms in BF16) and reaches its knee at 16K in both precisions (0.67 and 0.58 at 8K). Past the knee, the overlap ratio stays about 0.99 or above.

For each load $n$ we also restart the engine with $n_\text{ref}{=}n$ and plot the overlap ratio its calibration pass reports, $t_c/t_e$ (dashed). This estimate is conservative by construction: the profile pass skips attention and thus underestimates compute. Below the knee it lies under the measured ratio at every point but one (FP8 hybrid at 2K: 0.24 vs.\ 0.21), by up to 0.19; past the knee it stays at or above 0.94. A lower estimate only raises $t_e/t_c$ and hence $T$. At the per-step budget used in our evaluation ($n_\text{ref}{=}128$K), every configuration measures $t_e/t_c = 0.99$--$1.02$, so $T \approx \gamma \cdot C_\text{dummy}$, i.e., $n_T = \gamma \cdot (t_e/t_c) \cdot n_\text{ref} = 152$K--157K tokens, far beyond the measured knees (8K and 16K), so the calibrated $T$ keeps the transfers hidden behind compute.

H100 shares H200's FP8 and BF16 peak compute and 900~GB/s NVLink, the two quantities that set $n_T$ (Eq.~\eqref{eq:token_threshold}), so its thresholds match H200's; its 80~GB HBM only forces BF16 into hybrid mode, as in our A100/H100 BF16 runs.

\subsection{Compute Efficiency Around the Threshold}
\label{app:envelope_sweep}

\begin{figure}[!t]
\centering
\includegraphics[width=\linewidth]{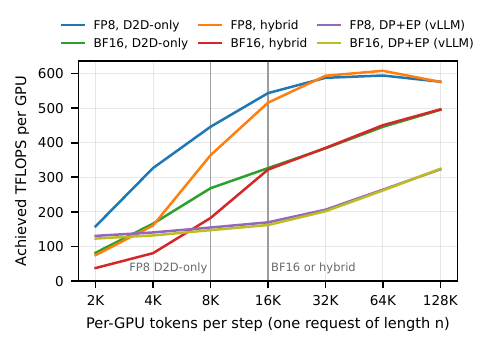}
\caption{Achieved FLOP rate per GPU (forward FLOPs over forward time) versus per-GPU tokens per step for D2D-only and hybrid-offloaded \SysBackend\ on Qwen3-235B-A22B (8$\times$H200, FP8 and BF16), with one request of length $n$ per step. Vertical gray lines mark the knees (overlap ratio ${\geq}0.99$; Fig.~\ref{fig:calibration_accuracy}): 8K for FP8 D2D-only and 16K for BF16 or hybrid.}
\label{fig:envelope_sweep}
\end{figure}

Figure~\ref{fig:envelope_sweep} reports the achieved FLOP rate per GPU (forward FLOPs over forward time) for the same sweep, with the knees marked. We plot this rate rather than throughput, which in any serving system rises with the tokens per step at light load and falls as attention's share of FLOPs grows with context, masking how well transfers are hidden. Below the knee, nothing switches mode: the router does not wait for more load (Appendix~\ref{app:impl_frontend}) and the backend does not fall back to \texttt{AllToAll} (Appendix~\ref{app:impl_backend}). Per-layer time, $\max(\text{compute}, \text{transfer})$ (\S\ref{subsec:async_ep_d2d}), stays pinned at the transfer time, e.g., 6.1--6.4~ms for FP8 D2D-only at 2K and 4K input tokens per step. With the step time flat, doubling the load from 2K to 4K input tokens roughly doubles both the achieved rate (158 to 327~TFLOPS) and throughput (26.2K to 51.4K tokens/s); at 8K input tokens, the FP8 D2D-only knee, compute catches up with the transfer (10.6~ms per layer) and the rate and throughput reach 446~TFLOPS and 63.5K tokens/s. Below the knee, transfers are partially exposed, but performance does not collapse. Past their knees, the two paths (D2D-only and hybrid) of each precision coincide, so offloading saves HBM without costing compute efficiency. vLLM DP+EP, measured the same way with the default setting, stays at 131--323~TFLOPS in FP8 and 123--325~TFLOPS in BF16 because its \texttt{AllToAll} remains on the critical path: past the knees \SysBackend\ achieves 1.8--3.2$\times$ (FP8) and 1.5--2.0$\times$ (BF16) its rate, while at 2K DP+EP is ahead of every path but FP8 D2D-only.
Single-request MFU is lower than in Fig.~\ref{fig:MFU_comparison_h100_random} at short and medium contexts because each of the $E{=}128$ experts receives only $nk/E$ tokens per step (1K at $n{=}16$K), too few to saturate its GEMMs; at 128K, FP8 MFU (29\%) matches Fig.~\ref{fig:MFU_comparison_h100_random}.

\section{Resource Costs of \SysBackend}
\label{app:costs}

\subsection{Per-Layer Communication Accounting}
\label{app:comm_accounting}

Table~\ref{tab:moe_comm_compare} lists on-critical-path traffic; Table~\ref{tab:comm_accounting} adds total physical traffic, counted as bytes \emph{received} per GPU per MoE layer. Under DP+EP, dispatch and combine deliver $2k\frac{P-1}{P}BSH \cdot b_a$ bytes, all on the critical path (vLLM's \texttt{AllGather}+\texttt{ReduceScatter} delivers $2(P{-}1)BSH \cdot b_a$, the same at $P{=}k{=}8$). These activations travel in BF16 ($b_a{=}2$) regardless of weight precision, so DP+EP moves the same bytes for FP8 and BF16 weights. \SysBackend's \texttt{AllGather} delivers $\frac{P-1}{P}E \cdot 3Hh \cdot b_w$ bytes regardless of $B{\cdot}S$, completely off the critical path once the load exceeds $n_T$. 

\begin{table}[!ht]
\centering
\footnotesize
\setlength{\tabcolsep}{4pt}
\begin{tabular}{@{}lrrrc@{}}
\toprule
\textbf{Strategy} & \textbf{$B{\cdot}S$/GPU} & \textbf{Analytical} & \textbf{Measured} & \textbf{On path} \\
\midrule
DP+EP,                  & 2K  & 0.22 & 0.23 & \checkmark \\
any weight precision    & 8K  & 0.88 & 0.89 & \checkmark \\
                          & 16K & 1.75 & 1.78 & \checkmark \\
                          & 32K & 3.50 & 3.56 & \checkmark \\
                          & 64K & 7.00 & 7.11 & \checkmark \\
\midrule
\SysBackend, FP8 weights  & any & 1.97 & 1.97 & -- \\
\SysBackend, BF16 weights & any & 3.94 & 3.94 & -- \\
\bottomrule
\end{tabular}
\caption{Bytes received per GPU per MoE layer (GiB) for Qwen3-235B-A22B on 8 GPUs; measured with NVLink hardware counters on 8$\times$H200. DP+EP moves BF16 activations whatever the weight precision, so its volume is the same for FP8 and BF16 weights. \SysBackend's \texttt{AllGather} is off the critical path when the load exceeds $n_T$.}
\label{tab:comm_accounting}
\end{table}

Measurements match both formulas to within 3\% (Table~\ref{tab:comm_accounting}). Because the DP+EP volume grows with the per-GPU load $B{\cdot}S$ while the \SysBackend\ volume is fixed, equating the two gives the load at which they cross:
\begin{equation}
  (BS)^{\ast} \;=\; \frac{3\,E\,h\,b_w}{2\,k\,b_a}.
  \label{eq:bs_crossover}
\end{equation}
The factor $\frac{P-1}{P}$ cancels, so $(BS)^{\ast}$ is independent of $P$. For Qwen3-235B-A22B, Eq.~\eqref{eq:bs_crossover} gives 18K tokens per GPU with FP8 weights and 36K with BF16 weights, both above the knees of Appendix~\ref{app:calibrated_T}. Between a knee and $(BS)^{\ast}$, \SysBackend\ moves \emph{more} bytes than DP+EP, yet all of them stay off the critical path, leaving no communication exposed; beyond $(BS)^{\ast}$, it moves even \emph{fewer} bytes, still entirely off the critical path. Its benefit therefore comes from removing communication from the critical path, not from reducing bytes (\S\ref{sec:analysis_system}).

\noindent\textbf{Transfer contention.} Replaying one layer's transfers on all eight GPUs, running the H2D copy concurrently with the \texttt{AllGather} leaves PCIe bandwidth unchanged (27.9~GB/s per GPU) and lowers \texttt{AllGather} bandwidth by 6\% in FP8 and 9\% in BF16.

\subsection{HBM Footprint of the Slot Ring}
\label{app:hbm}

Table~\ref{tab:hbm} breaks down per-GPU weight memory by execution mode; measured values on 8$\times$H200 match it within 0.1~GiB. The slot ring is allocated once at model load, before vLLM sizes the KV cache (Appendix~\ref{app:impl_backend}), so it takes a fixed share of HBM and never grows during serving. Its expert part, $S \cdot E \cdot 3Hh \cdot b_w$ bytes (11.25~GiB in FP8 and 22.5~GiB in BF16 at $S{=}5$), is \SysBackend's dedicated overhead and shrinks with $S$, down to 4.5~GiB in FP8 at the minimum $S{=}2$ in D2D-only mode (Appendix~\ref{app:impl_backend}), at the cost of less prefetch slack. For clarity, \S\ref{subsec:async_ep_offload} keeps attention weights resident; our implementation instead streams each layer's full weights, attention included, through the H2D slots in single-GPU and hybrid modes (Table~\ref{tab:hbm}, Attn.\ column). This frees 6.3~GiB per GPU in FP8 (12.6~GiB in BF16) for 3\% more PCIe bytes per layer on a single GPU. Keeping attention resident instead would leave every single-GPU and hybrid footprint in Table~\ref{tab:hbm} below 80~GB (at most 49.3~GiB), so the feasibility claims of \S\ref{subsec:eval_mfu} hold under either choice.

\begin{table}[!ht]
\centering
\footnotesize
\setlength{\tabcolsep}{2.5pt}
\begin{tabular}{@{}llrrrrrr@{}}
\toprule
 & \textbf{Mode ($P$)} & \textbf{Emb.} & \textbf{Attn.} & \textbf{Shards} & \textbf{H2D} & \textbf{Gather} & \textbf{Total} \\
\midrule
\multirow{4}{*}{FP8}  & Single GPU & 2.32 & --    & --    & 11.58 & --    & 13.90 \\
                      & Hybrid (2)   & 2.32 & --    & --    & 5.96  & 11.25 & 19.53 \\
                      & Hybrid (8)   & 2.32 & --    & --    & 1.74  & 11.25 & 15.31 \\
                      & D2D-only (8) & 2.32 & 6.29  & 26.44 & --    & 11.25 & 46.29 \\
\midrule
\multirow{4}{*}{BF16} & Single GPU & 2.32 & --    & --    & 23.17 & --    & 25.49 \\
                      & Hybrid (2)   & 2.32 & --    & --    & 11.92 & 22.50 & 36.74 \\
                      & Hybrid (8)   & 2.32 & --    & --    & 3.48  & 22.50 & 28.30 \\
                      & D2D-only (8) & 2.32 & 12.58 & 52.88 & --    & 22.50 & 90.27 \\
\bottomrule
\end{tabular}
\caption{Per-GPU weight memory (GiB) of Qwen3-235B-A22B at $S{=}5$. $P$: parallel degree (number of GPUs). \textbf{Emb.}: embedding and LM head (BF16). \textbf{Attn.}/\textbf{Shards}: resident attention weights and $\frac{1}{P}$ expert shards of all layers. \textbf{H2D} and \textbf{Gather}: the two parts of the $S$ shared slots, holding each layer's local weights (filled over PCIe) and its full expert set (assembled over NVLink by \texttt{AllGather}), respectively. KV cache and activations use the remaining HBM. The H2D slots hold $\frac{1}{P}$ of each layer's experts, so the hybrid footprint falls as $P$ grows; single-GPU mode streams whole layers and needs no Gather slots.}
\label{tab:hbm}
\end{table}

\begin{figure*}[t]
    \centering
    \includegraphics[width=\linewidth]{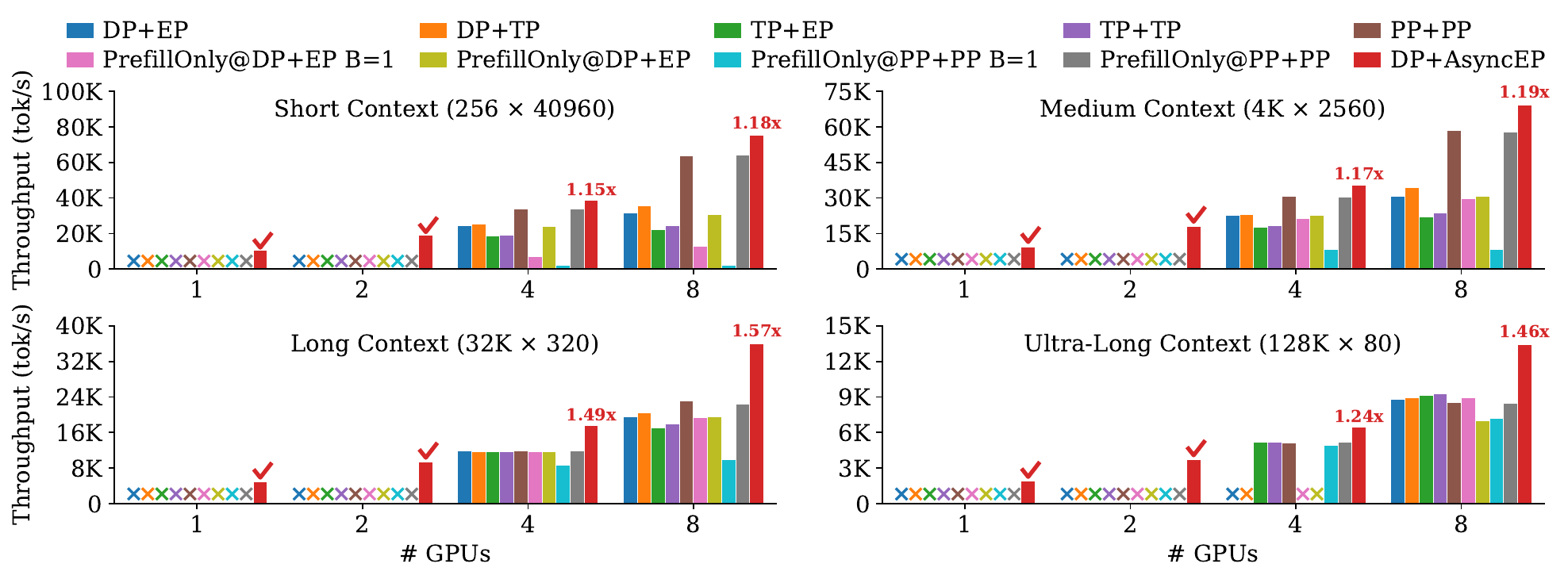}
    \caption{Throughput on GLM-4.5-FP8 under synthetic workloads with no prefix reuse, across four context regimes on 1--8 H200 GPUs. \textit{The four panels share the same total token count $S{\times}N$; DP+\SysBackend\ is \SysName's backend without prefix-aware routing. $\times$ marks configurations that run out of memory; annotations give DP+\SysBackend's gain over the strongest baseline at 4 and 8 GPUs.}}
    \label{fig:throughput_comparison_h200_random_glm}
\end{figure*}

\begin{figure*}[t]
  \centering
  \includegraphics[width=\linewidth]{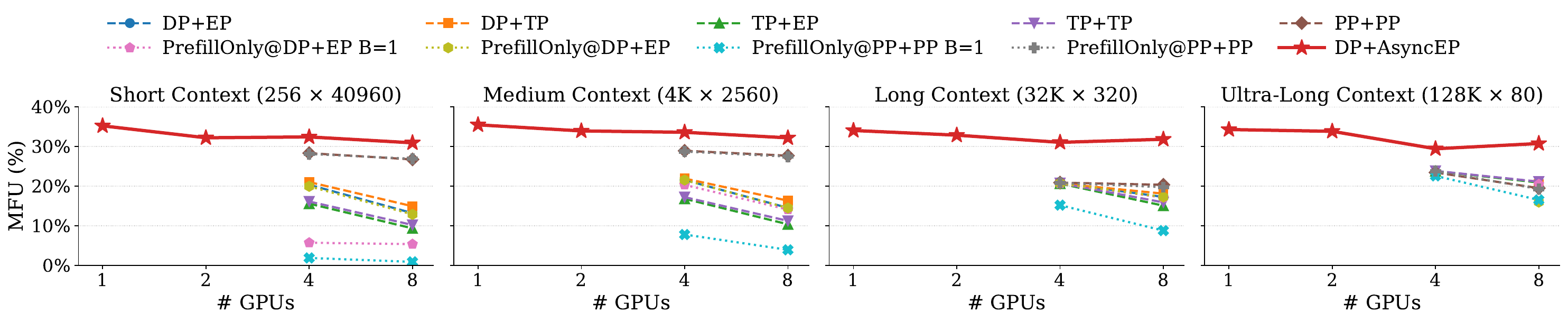}
  \caption{MFU on GLM-4.5-FP8 for the workloads of Figure~\ref{fig:throughput_comparison_h200_random_glm} at parallel degrees 1/2/4/8 on H200. \textit{Baselines have no points at 1--2 GPUs because the model does not fit; only DP+\SysBackend\ runs there.}}
  \label{fig:MFU_comparison_h200_random_glm}
\end{figure*}

\noindent\textbf{Effect on KV capacity.} With a 128K context on 8$\times$H200, vLLM's DP+EP leaves 62.3~GiB per GPU for the KV cache of Qwen3-235B-A22B in FP8. \SysName\ ($S{=}5$) matches it in D2D-only mode (62.4~GiB) and raises it to 93.3~GiB (1.5$\times$) in hybrid mode. In BF16, DP+EP leaves only 43.1~GiB, whereas hybrid mode leaves 86.5~GiB (2.0$\times$); D2D-only leaves 24.7~GiB, the one case where the slots cost KV capacity, yet still enough for a full 128K context. The extra HBM holds more cached prefixes, which the frontend's prefix-affinity routing (\S\ref{sec:frontend}) can turn into more prefix hits.

\section{Generality of \SysBackend}
\label{app:generality}

\subsection{Additional MoE Model}
\label{app:new_model}

We repeat the synthetic sweep of \S\ref{subsec:eval_generalization} on GLM-4.5-FP8~\cite{glm45}, an MoE model from a different family (355B parameters, 92 layers, 160 routed experts plus one shared expert, top-8 routing), on 1--8 H200 GPUs against the baselines of \S\ref{subsec:eval_setup}. As on Qwen3-235B-A22B, only DP+\SysBackend\ runs on 1--2 GPUs (Figure~\ref{fig:throughput_comparison_h200_random_glm}). At 8 GPUs it outperforms the strongest baseline by 1.18$\times$ (short), 1.19$\times$ (medium), 1.57$\times$ (long), and 1.46$\times$ (ultra-long); as in \S\ref{subsec:eval_generalization}, the gain is largest at long and ultra-long contexts, where on-path \texttt{AllToAll} volume and pipeline bubbles from unequal-cost prefill chunks penalize the baselines most.

Figure~\ref{fig:MFU_comparison_h200_random_glm} shows that DP+\SysBackend\ sustains 29.4--35.5\% MFU across all contexts and parallel degrees, whereas every baseline loses MFU from 4 to 8 GPUs, most sharply at short and medium contexts. \SysBackend's gains thus carry over to a second MoE family, as expected: the overheads it removes, on-path collectives and shrinking per-device batches, stem from distributed MoE execution rather than from a particular model (\S\ref{sec:analysis_system}).

\subsection{What Sets the Threshold Across MoE Families}
\label{app:threshold_families}
Expressing Eq.~\eqref{eq:threshold} in tokens per GPU per step makes its dependence on model and hardware explicit. Per layer, the \texttt{AllGather} moves $\frac{P-1}{P}E \cdot 3Hh \cdot b_w$ bytes at bandwidth $\beta$, while each token costs $2k \cdot 3Hh$ expert FLOPs plus $a$ additional FLOPs of attention and shared-expert work that needs no transfer. Hence
\begin{equation}
  n_T \;\approx\; \gamma \cdot \frac{P-1}{P} \cdot \frac{E}{k} \cdot \frac{b_w\,F_\text{ach}}{2\,\beta} \cdot \frac{1}{1 + a/(6kHh)}.
  \label{eq:token_threshold}
\end{equation}
Three consequences follow. (i)~Expert size $Hh$ cancels to first order: larger experts lengthen transfers and per-token compute alike, so they enlarge the absolute time slack but not the token load needed to hide transfers. (ii)~Routing sparsity $E/k$ matters instead: gpt-oss-120B~\cite{gptoss} and DeepSeek-V3~\cite{deepseekv3} ($E/k{=}32$) need about 2$\times$ more tokens per GPU than Qwen3-235B-A22B ($E/k{=}16$), partly offset by lower-precision weights (gpt-oss) or by a shared expert and heavier attention that enlarge $a$ (DeepSeek-V3). (iii)~Hardware and precision enter only through $b_w F_\text{ach}/\beta$. BF16 needs more tokens per GPU than FP8 (Appendix~\ref{app:calibrated_T}): it doubles $b_w$, while its achieved compute is only 1.2--1.6$\times$ lower. Across GPUs, in BF16 (A100 lacks FP8), H100 and H200 raise dense peak compute about 3$\times$ over A100 (989 vs.\ 312~TFLOPS) but NVLink bandwidth only 1.5$\times$ (900 vs.\ 600~GB/s), so they need about twice the tokens per GPU; A100 hides the same transfers with roughly half the load with \SysBackend.

\subsection{Extensions}
\label{app:discussion}

\noindent\textbf{Short decodes and speculative verification.}
\SysName\ targets prefill-only serving and is evaluated with one output token. For workloads that decode a few tokens after a long prefill, its mechanisms still apply to the prefill phase, e.g., as the prefill tier of a prefill/decode-disaggregated deployment~\cite{splitwise,distserve}; iterative decoding, whose per-step load falls far below $n_T$, remains outside the current design. A speculative-decoding verification pass~\cite{speculative} is a prefill-like forward over $B\cdot(d{+}1)$ tokens per GPU for draft length $d$, so \SysBackend\ applies once this load reaches $n_T$, as in throughput-oriented verification over large batches.

\noindent\textbf{Attention--expert disaggregation.}
Systems such as MegaScale-Infer~\cite{megascaleinfer} place attention and experts on separate GPU pools, ship activations between them, and partition experts statically across the expert pool. \SysBackend\ composes at the expert pool: expert GPUs gather each layer's weights so that any of them can serve any token, which removes routing-imbalance stragglers and lets attention GPUs dispatch by load rather than by expert ownership. The attention-to-expert activation transfer is inherent to disaggregation and stays on the critical path. The feasibility region follows from Appendices~\ref{app:envelope} and~\ref{app:hbm}: each expert GPU must receive at least $n_T$ tokens per step and host the $S$ full-layer expert slots. Under constrained resources, using fewer GPUs for the expert pool raises the per-GPU load, which eases overlap but tightens HBM; lowering $S$ toward its minimum (Appendix~\ref{app:impl_backend}) trades prefetch depth for capacity, and hybrid offloading (\S\ref{subsec:async_ep_offload}) keeps the expert shards in host memory, trading host memory for HBM at a higher $n_T$ (Appendix~\ref{app:calibrated_T}). Compressing these host-resident shards~\cite{su2024elves, zipllm_nsdi26, tensordex_sosp26} would further cut the host-memory cost.

\subsection{Serving Prefill-Only Decision Models}
\label{app:jev}

Prefill-as-a-service is quickly becoming a first-class product category, validating the central premise of this work. Jev~\cite{jev}, a recently announced \emph{decision model}, productizes the prefill-as-a-service interface of \S\ref{subsection:prefill-only_in_practice}: a request carries a state (text or JSON) and a set of typed questions over caller-defined options, and the model evaluates every question against the state in parallel, returning a probability for each option in a single request~\cite{jevdocs}. Its pricing makes the economics explicit: only input tokens are billed, and output tokens are free. With no text to generate, serving cost is governed by prefill throughput per GPU---exactly the metric \SysName\ is built to maximize.

\SysName\ is a natural serving substrate for decision models of this kind. (i)~\emph{Scale.} As such models grow toward frontier capability, MoE backbones that exceed one GPU's memory are the prevailing path (\S\ref{subsec:bg_moe}); \SysBackend\ removes per-layer \texttt{AllToAll} from their critical path and, with hybrid offloading, lets the same model run on as few as one GPU (\S\ref{subsec:eval_mfu}). On Qwen3-235B-A22B this yields 1.34--1.37$\times$ the throughput of the strongest baseline and 29.8--36.2\% MFU (\S\ref{sec:evaluation}). (ii)~\emph{Workflows.} TypeSafe's documentation recommends decomposing a complex judgment into many atomic questions, each evaluated in isolation against the same state~\cite{jevdocs}; served as prefill-only requests, these questions share the state as a long common prefix, which \SysName's frontend co-locates so that the shared state is prefilled once per GPU (\S\ref{subsec:prefix_routing}). (iii)~\emph{High-cardinality choices.} A choice question accepts up to 255 options~\cite{jevdocs}, and larger candidate sets are scored independently before a final selection~\cite{jev}, mirroring the multi-selection decomposition of \S\ref{subsection:prefill-only_in_practice}. (iv)~\emph{Seamless adoption.} \SysName\ requires no changes to model definitions or kernels and is enabled by a single flag on top of vLLM (Appendix~\ref{app:implementation}), so a decision model whose MoE backbone vLLM already supports can adopt it directly as its serving layer. Purpose-trained decision models and \SysName\ are thus two halves of one stack: the former define \emph{what} a prefill-only decision is, and \SysName\ makes executing it at scale fast and cheap. As prefill-only decision models proliferate, every advance on the model side multiplies the value of an efficient prefill-only serving layer, which positions \SysName\ as foundational infrastructure for this emerging class of workloads.

\section{Accuracy on a Production Classification Harness}
\label{app:accuracy_prod_harness}

This appendix details the Tier-2 evidence of \S\ref{subsec:eval_accuracy}: prefill-only reformulation preserves task accuracy relative to autoregressive (AR) greedy decoding. Because the claim is about the \emph{reformulation}, not the serving engine, we validate it on the nine-task classification harness used by our industry partners in production, run on a single NVIDIA H200 with two models---gpt-oss-120B~\cite{gptoss} (sparse MoE, 128 experts, MXFP4) for prefill-only and Llama-3.3-70B-Instruct-FP8~\cite{llama3} (dense) for AR, matching the partners' deployment split. Prompts are identically formatted; prefill-only scores each candidate by the log-probability of its canned answer token, realizing the atomic prefill operation of \S\ref{sec:analysis_workload}. Because the two protocols are attached to different models, Table~\ref{tab:accuracy_prod_harness} measures the end-to-end production delta (decoding mode + model capability); the same-model IMDB ablation (Table~\ref{tab:accuracy_imdb_ablation}) isolates the decoding-mode contribution.

\begin{table}[!ht]
\centering
\footnotesize
\setlength{\tabcolsep}{3pt}
\renewcommand{\arraystretch}{1.15}
\begin{tabular}{@{}llrrrrr@{}}
\toprule
\textbf{Dataset} & \textbf{Task} & \textbf{N} & \textbf{K} & \textbf{AR} & \textbf{Pref.} & \textbf{$\Delta$} \\
\midrule
EXACT\_MATCH  & Entity mention    &    510 &  2 & 99.0 & \textbf{99.2} & $+0.2$ \\
Simple\_Facts & Fact verify       &    500 &  2 & \textbf{96.8} & 95.2 & $-1.6$ \\
IMDB~\cite{imdb}          & Sentiment         & 25{,}000 &  2 & \textbf{95.1} & 94.1 & $-1.0$ \\
Toxicity~\cite{toxicity_surge}      & Toxicity          & 1{,}000 &  2 & \textbf{80.1} & 78.7 & $-1.4$ \\
Twitter-Fin~\cite{twitterfin}   & 3-way sentiment   & 1{,}000 &  3 & 71.9 & \textbf{83.6} & $+11.7$ \\
INTENTS       & Intent classify   &    328 & 10 & \textbf{97.9} & 90.9 & $-7.0$ \\
RoGuard       & Content policy    &    999 & 23 & \textbf{59.1} & 55.5 & $-3.6$ \\
Unfair-ToS~\cite{lexglue}    & Legal clause      & 1{,}000 & 32 & 38.6 & \textbf{40.0} & $+1.4$ \\
GoEmotions~\cite{goemotions}    & Fine-grain emotion & 1{,}000 & 28 & 35.6 & \textbf{36.3} & $+0.7$ \\
\bottomrule
\end{tabular}
\caption{Task accuracy (\%) on nine production benchmarks. \textbf{N}: \#rows. \textbf{K}: \#candidate labels. \textbf{AR}: Llama-3.3-70B-Instruct-FP8 (greedy). \textbf{Pref.}: gpt-oss-120B (prefill-only logit scoring). Bold marks the better value per row.}
\label{tab:accuracy_prod_harness}
\end{table}

\noindent\textbf{Cross-task observations.}
The results split into a \emph{decoding-invariant band}---seven of nine tasks within $\pm 3.6$\,pp of AR across label cardinalities from 2 to 32, with prefill-only winning 3 and AR winning 4, consistent with sample-level variance at these dataset sizes---and two \emph{model-specific outliers} (Twitter-Fin $+11.7$\,pp, INTENTS $-7.0$\,pp), both attributable to prompt-design sensitivity of the underlying model rather than to the decoding mode.

\begin{table}[!ht]
\centering
\footnotesize
\setlength{\tabcolsep}{3pt}
\renewcommand{\arraystretch}{1.15}
\begin{tabular}{@{}lrrr@{}}
\toprule
\textbf{Decoding mode}                                & \textbf{Acc.} & \textbf{Tok/s} & \textbf{Time} \\
\midrule
Autoregressive~+ full chain-of-thought                & 95.1 & 3{,}525  & 3{,}215\,s \\
Autoregressive, final-channel only (stub)             & 94.1 & 22{,}508 & 410\,s     \\
Prefill-only (logit scoring, no generation)           & 91.8 & 22{,}619 & 828\,s     \\
\bottomrule
\end{tabular}
\caption{IMDB three-mode ablation on gpt-oss-120B, 25K rows, 1$\times$H200. Rows~2 vs.~3 isolate the decoding-mode effect under a fixed prompt template.}
\label{tab:accuracy_imdb_ablation}
\end{table}

\noindent\textbf{Isolating the decoding-mode gap.}
The 3.3-pp total gap between full-reasoning AR and prefill-only decomposes into 1.0\,pp from removing the chain-of-thought channel (rows~1$\to$2; a prompt-design choice not required by our reformulation) and 2.3\,pp from switching token generation to logit scoring (rows~2$\to$3). Under a fixed prompt template, prefill-only therefore costs 2.3\,pp accuracy while delivering 6.4$\times$ higher tokens/s than full-reasoning AR.

\noindent\textbf{Anchor for Tier~1.}
The prefill-only row of Table~\ref{tab:accuracy_imdb_ablation} is produced on \SysName's backend with weight offloading enabled, so 91.8\% is the accuracy attained \emph{on \SysName's execution stack}; the non-offloaded vLLM baseline on the same model, prompts, and scoring procedure lands within measurement noise, consistent with the backend mechanisms of \S\ref{sec:backend} preserving model outputs up to FP non-associativity (Tier~1 of \S\ref{subsec:eval_accuracy}).

\end{document}